\documentclass[runningheads]{llncs}

 \usepackage[dvipsnames,svgnames,table]{xcolor} 
 
\usepackage{eccv}

\usepackage{eccvabbrv}

\usepackage{graphicx}
\usepackage{booktabs}

\usepackage[accsupp]{axessibility}  

\usepackage{hyperref}

\usepackage{orcidlink}

\usepackage{url}
\usepackage{times}
\usepackage{epsfig}
\usepackage{amssymb}
\usepackage{amsmath} 
\usepackage{bm} 
\usepackage{subcaption}
\usepackage{graphbox}
\usepackage{paralist}
\usepackage{amsfonts}

\usepackage{makecell}
\usepackage{tabularx}
\usepackage{multirow}
\usepackage{pifont}
\usepackage[para,online,flushleft]{threeparttable}

\usepackage{algorithm}
\usepackage{setspace}
\usepackage[noend]{algpseudocode}

\usepackage{color}

\usepackage{stmaryrd} 

\usepackage[normalem]{ulem}

\def\mX{{\bm{X}}}
\def\mW{{\bm{W}}}
\def\mI{{\bm{I}}}
\def\gQ{{\mathcal{Q}}}
\def\gP{{\mathcal{P}}}
\def\vv{{\bm{v}}}

\def\Plus{\texttt{+}}
\def\Minus{\texttt{-}}
\def\Equal{\texttt{=}}

\begin{document}

\title{Overcoming Distribution Mismatch\texorpdfstring{\\}{} in Quantizing Image Super-Resolution Networks}

\titlerunning{ODM}

\author{Cheeun Hong\inst{1}\orcidlink{0009-0009-3480-748X} \and
Kyoung Mu Lee\inst{1,2}\orcidlink{0000-0001-7210-1036}
}

\authorrunning{C.~Hong and K.M.~Lee}


\institute{
Dept. of ECE \& ASRI, \email{\{cheeun914, kyoungmu\}@snu.ac.kr} \and
IPAI, Seoul National University
}

\maketitle
\begin{abstract}
Although quantization has emerged as a promising approach to reducing computational complexity across various high-level vision tasks, it inevitably leads to accuracy loss in image super-resolution (SR) networks.
This is due to the significantly divergent feature distributions across different channels and input images of the SR networks, which complicates the selection of a fixed quantization range. 
Existing works address this distribution mismatch problem by dynamically adapting quantization ranges to the varying distributions during test time.
However, such a dynamic adaptation incurs additional computational costs during inference.
In contrast, we propose a new quantization-aware training scheme that effectively \textbf{O}vercomes the \textbf{D}istribution \textbf{M}ismatch problem in SR networks without the need for dynamic adaptation.
Intuitively, this mismatch can be mitigated by regularizing the distance between the feature and a fixed quantization range.
However, we observe that such regularization can conflict with the reconstruction loss during training, negatively impacting SR accuracy.
Therefore, we opt to regularize the mismatch only when the gradients of the regularization are aligned with those of the reconstruction loss.
Additionally, we introduce a layer-wise weight clipping correction scheme to determine a more suitable quantization range for layer-wise weights.
Experimental results demonstrate that our framework effectively reduces the distribution mismatch and achieves state-of-the-art performance with minimal computational overhead.
Codes are available at \href{https://github.com/Cheeun/ODM}{https://github.com/Cheeun/ODM}.
\end{abstract}

\section{Introduction}
\label{sec:introduction}

Image super-resolution (SR) is a core low-level vision task aimed at reconstructing high-resolution (HR) images from their corresponding low-resolution (LR) counterparts.
Recent advances in deep learning~\cite{dong2015image,kim2016accurate,lim2017enhanced,zhang2018rcan,zhang2018residual,liang2021swinir,chen2023activating,park2023content,son2021toward} have led to astonishing achievements in producing high-fidelity images.
However, the remarkable performance is based on heavy network architectures that incur significant computational costs, limiting practical viability, such as mobile deployment.

To mitigate the computational complexity of neural networks, quantization has emerged as a promising avenue. 
Network quantization has proven effective in reducing computational costs without significant accuracy loss, particularly in high-level vision tasks, such as image classification~\cite{choi2018pact,hou2018loss,zhou2016dorefa}.
However, when it comes to quantizing SR networks to lower bit-widths, substantial performance degradation~\cite{ignatov2021real} often occurs, posing a persistent and challenging problem to be addressed.

\newcommand{\cm}{{\ding{51}}}%
\newcommand{\xm}{{\ding{55}}}%
\begin{table}[!t]
    \centering
    \setlength{\tabcolsep}{1.3mm}
    \caption{
    \textbf{Quantization methods on SR.} For performance, existing methods rely on channel-wise quantization or input-adaptive module, which incur computational overhead. Our method achieves high accuracy without utilizing channel-wise quantization or input-adaptive modules.
    } 
    \resizebox{0.65\linewidth}{!}{
            \begin{tabular}{l cc cc}
                \toprule
                Method & Channel-wise Q & Input-adaptive Modules & PSNR & SSIM \\
                \midrule
                PAMS~\cite{Li2020pams} & \xm & \xm & 29.51 & 0.835\\
                DDTB~\cite{zhong2022ddtb} & \xm & \cm & 30.97 & 0.876\\
                DAQ~\cite{hong2022daq} & \cm & \cm & 31.01 & 0.871\\
                \midrule
                Ours & \xm & \xm & \textbf{31.50} & \textbf{0.882} \\
                \bottomrule
            \end{tabular}
    }
    \label{tab:intro-first}
\end{table}

This degradation can be attributed to the significant variance in the feature (activation) distributions of SR networks.
The feature distribution of a layer exhibits substantial discrepancies across different channels and images, which makes it difficult to determine a single quantization range for a layer.
Early approaches to SR quantization~\cite{Li2020pams} adopt a training scheme to learn the quantization range of each layer.
However, despite careful selection, the quantization ranges do not align with the varied values within the channel and image dimensions, which we refer to as \textit{distribution mismatch} in features.

Recent approaches aim to address this issue by incorporating dynamic adaptation modules that accommodate the varying distributions.
For example, the quantization range is dynamically adjusted by directly leveraging the distribution mean and variance at test time~\cite{hong2022daq} or by employing input-adaptive dynamic modules~\cite{zhong2022ddtb}.
Although adapting the quantization function to each image during inference might handle variable distributions, the dynamic modules introduce considerable computational overhead, potentially compromising the computational benefits of quantization.

In this study, we propose a novel quantization-aware training framework that addresses the distribution mismatch problem with a loss term that regulates the mismatch in distributions.
Although directly minimizing the feature mismatch presents the potential for quantization-friendliness, whether it preserves reconstruction accuracy is questionable. 
We observe that concurrent optimization with the mismatch regularization and the original reconstruction loss can disrupt the image reconstruction process.
Therefore, we introduce a cooperative mismatch regularization strategy, where the mismatch is regulated only when it collaborates harmoniously with the reconstruction loss. 
To determine the cooperative behavior, we assess the cosine similarity of the gradients from each loss, then we weigh the gradients of mismatch regularization based on this similarity.
Consequently, we effectively update the SR network to hold both quantization-friendliness and reconstruction accuracy.

Furthermore, we identify the distribution mismatch among the weights of different layers.
We discover that employing a fixed policy to determine the layer-wise weight quantization range~\cite{Li2020pams,hong2022daq,hong2022cadyq,tian2023cabm} can be suboptimal and that a further precise range can be obtained by considering both the current distribution of weights and the distinct tendencies of each layer.
Therefore, we additionally incorporate layer-specific variations using a correction parameter for each layer.
This strategy allows us to accurately find the quantization range for weights while incurring only a minimal overhead (0.01\% additional storage size and no additional bitOPs)--a significantly smaller impact compared to methods that use dynamic modules.
Overall, the contributions of our work include the following:
\begin{itemize}
    \item[$\bullet$] We introduce the first quantization framework to address the distribution mismatch problem in SR networks without dynamic modules, as compared in \Cref{tab:intro-first}. Our framework updates the SR network to be quantization-friendly and accurate simultaneously.
    \item[$\bullet$] Based on the observations of the distribution mismatch in SR networks, we effectively reduce the mismatch by introducing a cooperative mismatch regularization term and a weight clipping correction term.
    \item[$\bullet$] Compared to existing approaches to SR quantization, ours achieves state-of-the-art performance with similar or fewer computations. 
\end{itemize}

\section{Related Works}
\label{sec:related_works}
\subsection{Image Super-Resolution}
Convolutional Neural Network (CNN)-based approaches have demonstrated remarkable advancements in the image super-resolution (SR) task~\cite{ledig2017photo,lim2017enhanced}, but at the cost of substantial computational resources.
The intensive computations required by SR networks have spurred interest in developing lightweight SR architectures~\cite{dong2016srcnn,hui2019imdn,hui2018idn,zhang2018rcan,jo2021practical}.
Furthermore, various strategies for lightweight SR networks have been explored, including neural architecture search~\cite{chu2021fast,kim2022fine,li2020dhp,song2020efficient,li2021heterogeneity}, knowledge distillation~\cite{hui2018idn,hui2019imdn,zhang2021data}, and pruning~\cite{jiang2021learning,Oh_2022_CVPR,zhang2022learning}.
While these methods predominantly focus on reducing network depth or the number of channels, our work specifically aims to lower the precision of floating-point operations through network quantization.

\subsection{Network Quantization} 
By mapping 32-bit floating point values of features and weights in convolutional layers to lower-bit representations, network quantization provides a dramatic reduction in computational resources~\cite{cai2017deep,choi2018pact,esser2019learned,jung2019learning,zhou2016dorefa,zhuang2018towards}.
Recent works have successfully quantized various networks to low bit-widths with minimal compromise in accuracy~\cite{cai2020rethinking,dong2019hawq,habi2020hmq,jin2020adabits,lou2019autoq,wang2019haq,yang2020fracbits}.
However, these efforts primarily target high-level vision tasks, whereas networks for low-level vision tasks remain vulnerable to low-bit quantization.

\subsection{Quantized Super-Resolution Networks} 
In contrast to high-level vision tasks, SR poses different challenges due to its inherent high sensitivity to quantization~\cite{ignatov2021real,ma2019efficient,xin2020binarized,Wang2021fully}.
Some works have attempted to recover accuracy by modifying the network architecture~\cite{ayazoglu2021extremely,jiang2021training,xin2020binarized}
or by assigning different bits for each image~\cite{hong2022cadyq,tian2023cabm,Hong_2024_CVPR} or network stage~\cite{liu2021super}.
However, the primary challenge in quantizing SR networks lies in the vastly distinct feature distributions.
To address this, Li \etal~\cite{Li2020pams} adopted a learnable quantization range for different layers.
Subsequently, recognizing that the distributions vary not only by layer, but also by channel and input, Hong \etal~\cite{hong2022daq} introduced a channel-wise dynamic quantization function.
Additionally, Zhong \etal~\cite{zhong2022ddtb} utilized an input-adaptive dynamic module to tailor quantization ranges for each specific input image.
However, such dynamic adaptations of quantization functions during test-time incur non-negligible computational overheads.
Instead of relying on test-time adaptive modules, our approach focuses on mitigating the feature mismatch before quantization.
Our framework reduces the inherent distribution mismatch in SR networks with minimal overhead, accurately quantizing SR networks without dynamic modules.
More recently, Qin \etal~\cite{qin2024quantsr} introduced additional transformation functions in both the forward and backward processes.
However, performance degradation is still evident in ultra-low bit (\eg, 2-bit) scenarios.
\section{Proposed Method}
\label{sec:proposed_method}
In this section, after a brief introduction to network quantization (\Cref{subsec:preliminaries}), we first analyze the mismatch in the features of SR networks (\Cref{subsec:activation-mismatch}).
Subsequently, we propose a solution to reduce this feature mismatch during training (\Cref{subsec:cooperative-mismatch}).
Additionally, we examine the mismatch in weights across SR networks (\Cref{subsec:weight-mismatch}) and introduce a quantization range selection scheme that addresses the weight mismatch (\Cref{subsec:weight-clipping}).
The overall training process is summarized in \Cref{algo-odm}.

\subsection{Preliminaries}
\label{subsec:preliminaries}
To reduce the heavy computations of convolutional and linear layers in neural networks, the input features (activations) and weights of each convolution/linear layer are quantized to low-bit values~\cite{cai2017deep,choi2018pact,jung2019learning,gholami2022survey}.
Given the input feature of the $i$-th convolution/linear layer $\mX_i\in\mathbb{R}^{B\times C\times H\times W}$, where $B, C, H,$ and $W$ represent the dimensions of the input batch, channel, height, and width, respectively, a quantization operator $q(\cdot)$ quantizes the feature $\mX_i$ with bit-width $b$:
\begin{equation}
\label{eq:quant}
    q(\mX_i; l, u) = \text{Int}({ \frac{\text{clip}(\mX_i, l, u) - l}{s} }) \cdot s + l,
\end{equation} 
where $ \text{clip}(\cdot, l, u)$ truncates the input into the quantization range of $[l, u]$ and $s = \frac{u - l}{2^b-1}$.
After truncation, the truncated feature is scaled to $[0, 2^b-1]$, then rounded to integer values with Int$(\cdot)$, and rescaled to range $[l, u]$.

For activation quantization, the quantization range is defined by [$l_a, u_a$].
To obtain better quantization ranges in SR networks, the clipping parameters $l_a, u_a$ for each layer are typically learned through quantization-aware training~\cite{Li2020pams,zhong2022ddtb}.
Since the rounding function is not differentiable, a straight-through estimator (STE)~\cite{bengio2013estimating} is used to train the clipping parameters in an end-to-end manner.
Following~\cite{zhong2022ddtb}, we initialize $l_a$ and $u_a$ as the (100-$j$)-th and $j$-th percentile values of the feature, averaged among the training data, where $j$ is set to 99 in our experiments to avoid outliers that corrupt the quantization range.
For weights, as their distributions tend to be symmetric and the mean approximates zero, we utilize a symmetric quantization function.
Thus, for the weight $\mW$ of each convolution/linear layer, the quantization range is defined by [$-u_w, u_w$].

\begin{figure}[t]
\centering
\includegraphics[width=0.98\textwidth]{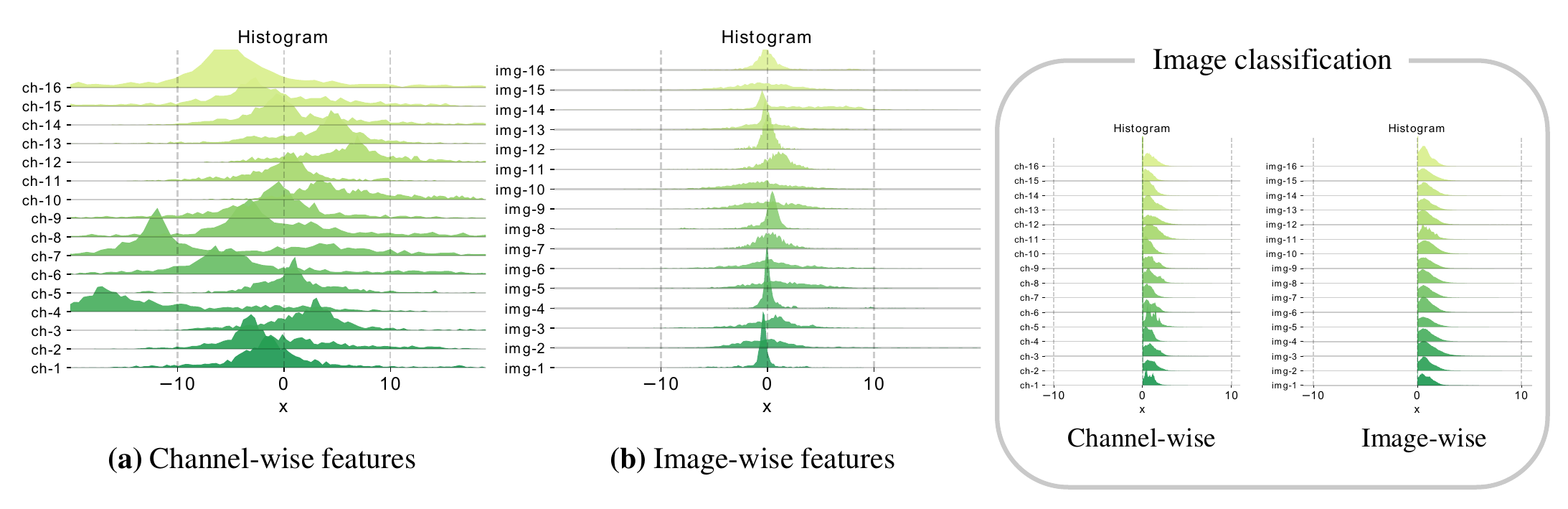}
\caption{
    \textbf{Distribution mismatch in SR networks.} 
    Compared to a classification network (\eg, ResNet-18), an SR network (\eg, EDSR) exhibits significant mismatches within the feature distributions across channel and image dimensions.
    The large distribution mismatch complicates the selection of an appropriate quantization range.
    Channels and images from the 2nd layer are randomly selected for visualization.
    Additional results are available in the supplementary material.
}
\label{fig:method-motiv}
\end{figure}

\subsection{Distribution Mismatch in SR Networks}
\label{subsec:activation-mismatch}

The unfriendliness to quantization in SR networks arises from diverse feature (activation) distributions, as reported in previous studies~\cite{Li2020pams,hong2022daq,zhong2022ddtb}, primarily due to the absence of batch normalization layers in SR networks.
As illustrated in \Cref{fig:method-motiv}, where there are notable discrepancies between the channel and image distributions, quantization grids are unnecessarily allocated to regions with minimal feature density.
Early SR quantization methods tackled this issue by employing learnable quantization range parameters~\cite{Li2020pams} for each feature.
However, even though the quantization-aware training process strives to find the optimal range for each feature, it fails to account for the channel-wise and input-wise variance in distributions.
This mismatch results in a large quantization error that can impair SR performance.
To deal with these distribution mismatches, existing methods adopt different quantization ranges for each channel~\cite{hong2022daq} or input image~\cite{zhong2022ddtb,hong2022daq}.
Nevertheless, the test-time adaptation modules used to determine these ranges introduce unwanted computational overhead during inference.
Thus, our straightforward solution is to pre-adjust the distributions to be quantization-friendly, thereby eliminating the need for additional adaptations at inference.
The following sections will introduce a new quantization-aware training scheme designed to resolve the distribution mismatch problem.

\subsection{Cooperative Mismatch Regularization}
\label{subsec:cooperative-mismatch}

\begin{figure}[t]
    \centering
    \begin{subfigure}{0.3\textwidth}
        \centering
        \includegraphics[width=\textwidth]{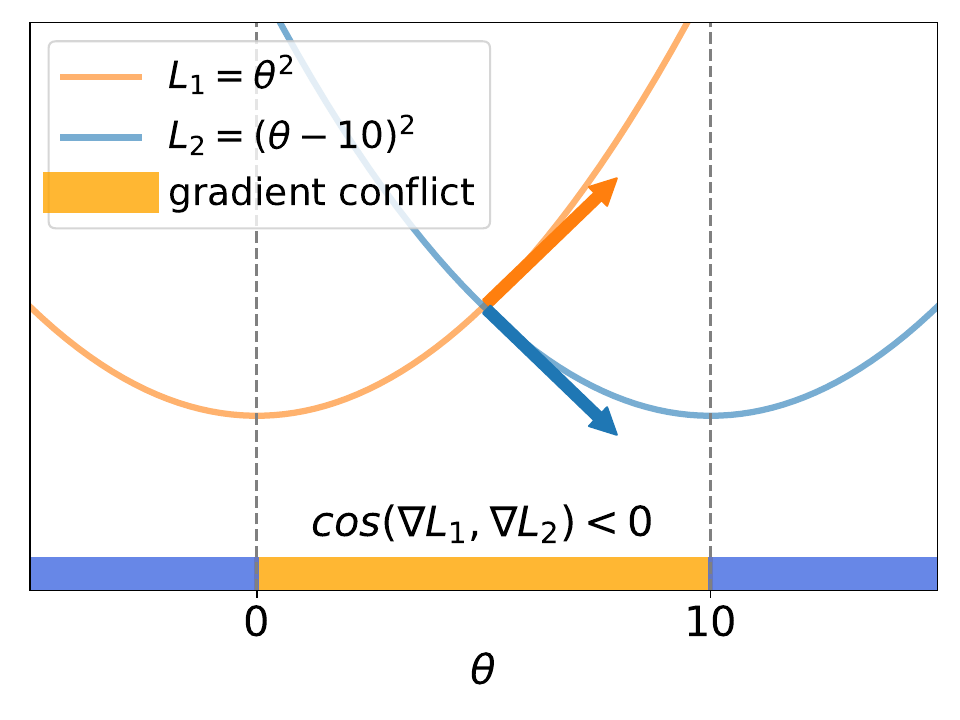}
        \caption{
        Gradient conflict~\cite{du2018adapting}
        \label{fig:method-conflict-a}}
    \end{subfigure}
    \begin{subfigure}{0.3\textwidth}
        \centering
        \includegraphics[width=\textwidth]{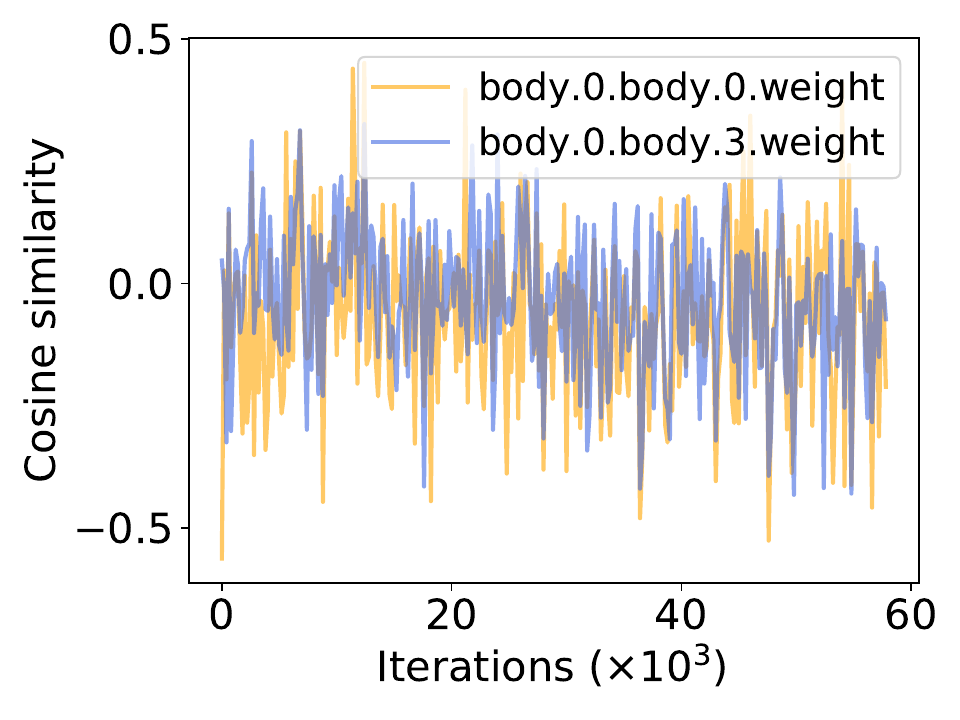}
        \caption{Cosine similarity of two gradients\label{fig:method-conflict-b}}
    \end{subfigure} 
    \begin{subfigure}{0.3\textwidth}
        \centering
        \includegraphics[width=\textwidth]{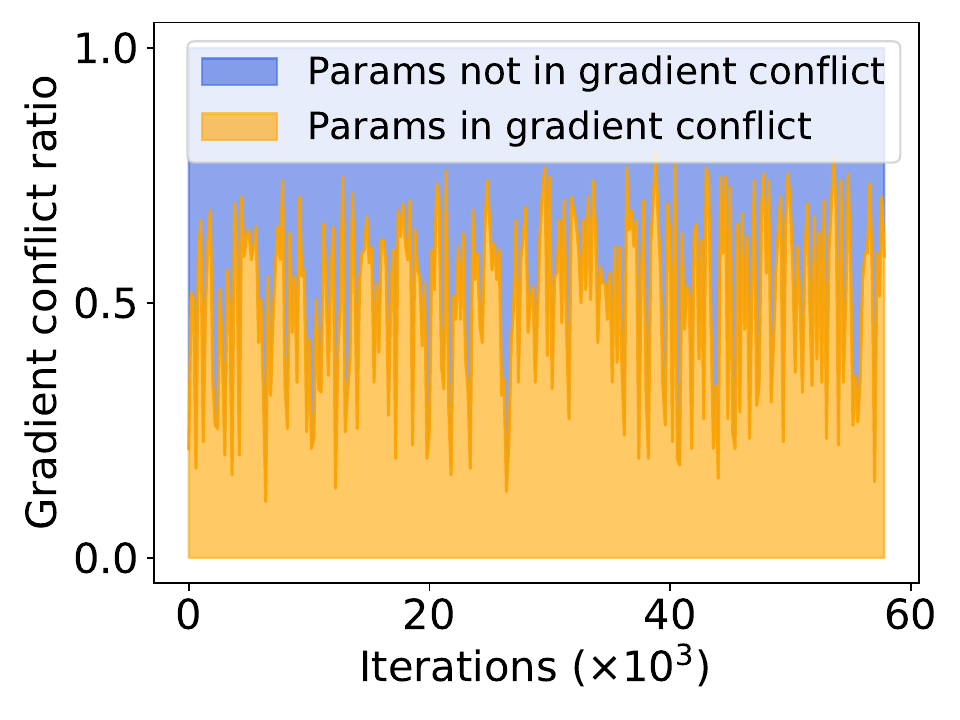}
        \caption{Gradient conflict ratio \label{fig:method-conflict-c}}
    \end{subfigure}
    \caption{ 
        \textbf{Conflict between mismatch regularization and reconstruction loss.} 
        Mismatch regularization updates a number of parameters in the \textit{contradictory} direction to the reconstruction loss, which we refer to as gradient conflict.
        (b) When the two losses are jointly used, gradient conflict consistently occurs during training, outputting a negative cosine similarity value.
        (c) We plot the ratio of conflicted gradients during training.
        Nearly half of the parameters undergo gradient conflict, which indicates that merely combining mismatch regularization with the reconstruction loss can impair SR accuracy. 
        Visualizations are done on EDSR.
    }
    \label{fig:method-conflict}
\end{figure}

Instead of trying to identify a quantization range capable of accommodating diverse feature distributions, our approach aims to regularize the distribution mismatch beforehand.
Obtaining an appropriate quantization range for a feature with high image- and channel-wise variance is difficult, as a certain number of channels or images will invariably be distant from the selected quantization range.
In this work, we refer to the total distance of each feature from the selected quantization grid as the mismatch,
\begin{equation}
    M(\mX_i) = ||\mX_i-q(\mX_i; l_a, u_a)||_2,
\label{eq:mismatch}
\end{equation}
where $||\cdot||_2$ calculates the Frobenius norm.
Further analyses of the definition of mismatch are provided in the supplementary material.
We can reduce the overall mismatch by directly regularizing the mismatch of each feature to be quantized.
The mismatch regularization loss is obtained by summing the mismatch over all quantized features:
\begin{equation}
    \mathcal{L}_{M} = \sum_i^{\#\text{layers}}  M(\mX_i).
\label{eq:loss-mismatch}
\end{equation}
The mismatch regularization loss can be used in line with the original reconstruction loss typically used in the general quantization-aware training pipeline for SR networks:
\begin{equation}
    \mathcal{L}_R = \mathcal{L}_1(\gQ(\mI_{LR}), \mI_{HR}), 
\label{eq:loss-rec}
\end{equation}
where
$\mathcal{L}_1$ loss indicates the $l_1$ distance between the reconstructed image using the quantized network $\gQ$ and the ground-truth HR image $\mI_{HR}$.
Then, the optimization of the parameter $\theta^t$ is formulated as:
\label{subsec:loss}
\begin{equation}
    {\theta}^{t+1} = \theta^{t} - \beta^{t} \cdot (\nabla_\theta \mathcal{L}_R(\theta^t) + \nabla_\theta \mathcal{L}_M(\theta^t) ), 
\end{equation}
where $\nabla_\theta \mathcal{L}_R(\theta^t)$ denotes the gradient from the original reconstruction loss and $\nabla_\theta \mathcal{L}_M(\theta^t)$ is the gradient from mismatch regularization loss, and $\beta^t$ refers to the learning rate. 
Updating the network to minimize the mismatch regularization loss will reduce the overall error from feature quantization.

However, then a question arises: \textit{does reducing the quantization error of each feature lead to improved reconstruction accuracy}?
The answer is, according to our observation in \Cref{fig:method-conflict}, not necessarily.
During the training process, the mismatch regularization loss can collide with the original reconstruction loss.
That is, for some parameters, the direction of the gradient from reconstruction loss and that of the mismatch regularization are opposing, referred to as gradient conflict~\cite{du2018adapting}.
As in \Cref{fig:method-conflict-b}, the cosine similarity of two gradients oscillates between positive and negative values during training, indicating that the directions of two gradients do not converge and the gradient occasionally conflicts.
Furthermore, as shown in \Cref{fig:method-conflict-c}, the proportion of parameters undergoing gradient conflict is not minor, implying that the regularization loss can severely hinder the reconstruction loss.

We aim to avoid the conflict between these two losses, in other words, to minimize the mismatch as long as it does not hinder the reconstruction loss.
Thus, we dismiss the mismatch regularization term when it is not cooperative with reconstruction loss and make more use of it when it is cooperative.
Specifically, we determine whether the two losses are cooperative by examining the cosine similarity of the gradients of each loss and then simply weigh the gradient of mismatch regularization by the gradient similarity. 
Our cooperative mismatch regularization can be formulated as follows:
\begin{equation}
    \theta^{t+1} = 
    \theta^{t} - \beta^{t} \cdot (\lambda_R \cdot \nabla_\theta \mathcal{L}_R(\theta^t) + \lambda_M \cdot \underline{sim(\nabla_\theta \mathcal{L}_R(\theta^t), \nabla_\theta \mathcal{L}_M(\theta^t))} \cdot \nabla_\theta \mathcal{L}_M(\theta^t) ),    
    \label{eq:loss-coop}    
\end{equation}
where the underlined term, gradient similarity, is defined as $sim(\vv_a, \vv_b)\Equal \frac{cos(\vv_a, \vv_b)\Plus 1}{2}$ and $cos(\cdot, \cdot)\in[\Minus1,1]$ calculates the cosine similarity between two vectors. $\lambda_R, \lambda_M$ are hyper-parameters to balance the two gradients.
If the directions of two gradients are similar (\ie, smaller than $90^\circ$ and closer to $0^\circ$), the gradient similarity is a large value, then the parameter is updated in the direction where mismatch regularization is also substantially considered.
On the contrary, if the two gradients point in the other direction (\ie, larger than $90^\circ$ and closer to $180^\circ$), the two losses restrain each other, and the gradient similarity is close to 0.
In this case, we follow the gradient of reconstruction loss.
This allows the network to reduce the quantization error cooperatively with the reconstruction error.
Details on gradient similarity are in the supplementary material.

\subsection{Distribution Mismatch in Weights}
\label{subsec:weight-mismatch}
\begin{figure}[t]
    \centering
    \begin{subfigure}{0.3\textwidth}
        \centering
        \setlength{\tabcolsep}{1.2mm}
        \includegraphics[width=0.95\textwidth]{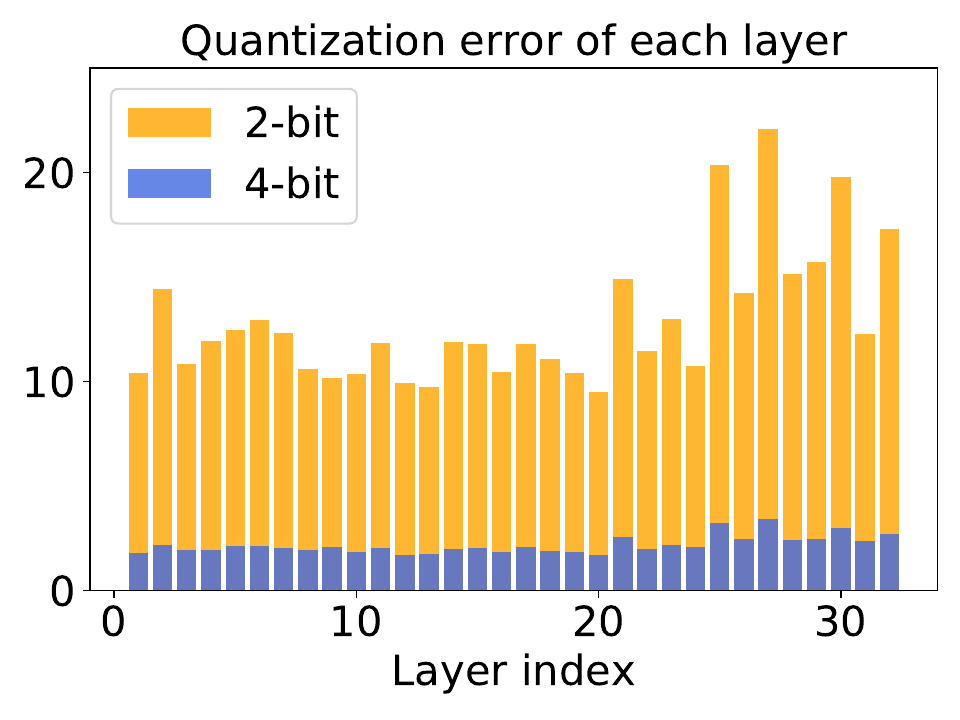}
        \caption{Layer-wise QE using max policy\label{fig:method-weight-a}}
    \end{subfigure}
    \begin{subfigure}{0.3\textwidth}
        \centering
        \setlength{\tabcolsep}{1.2mm}
        \includegraphics[width=0.95\textwidth]{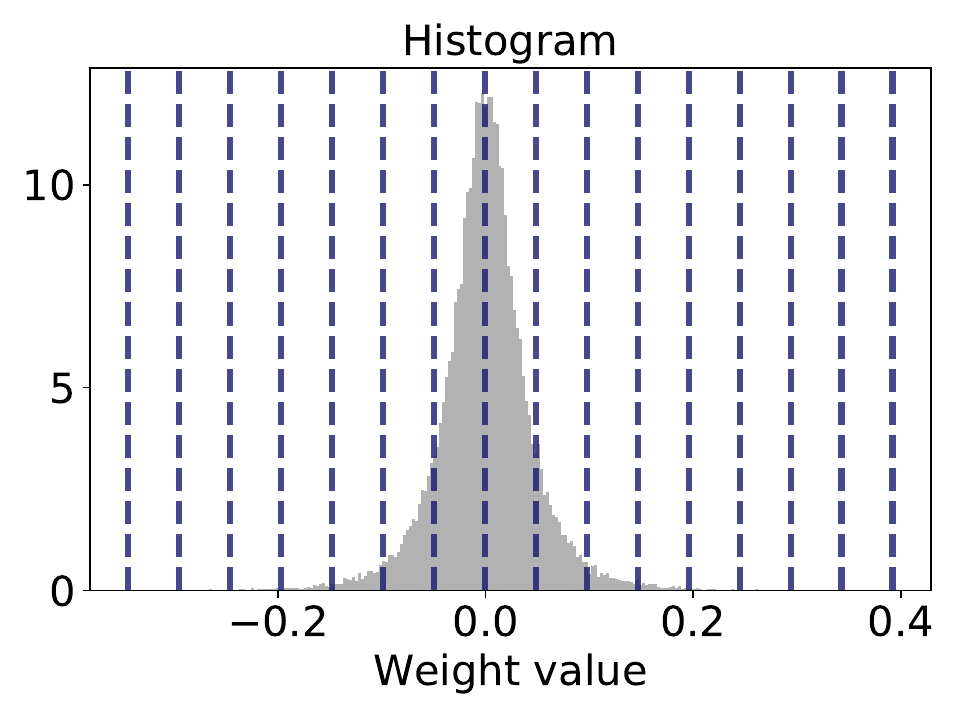}
        \caption{4-bit quantization grid\label{fig:method-weight-b}}
    \end{subfigure} 
    \begin{subfigure}{0.3\textwidth}
        \centering
        \setlength{\tabcolsep}{1.2mm}
        \includegraphics[width=0.95\textwidth]{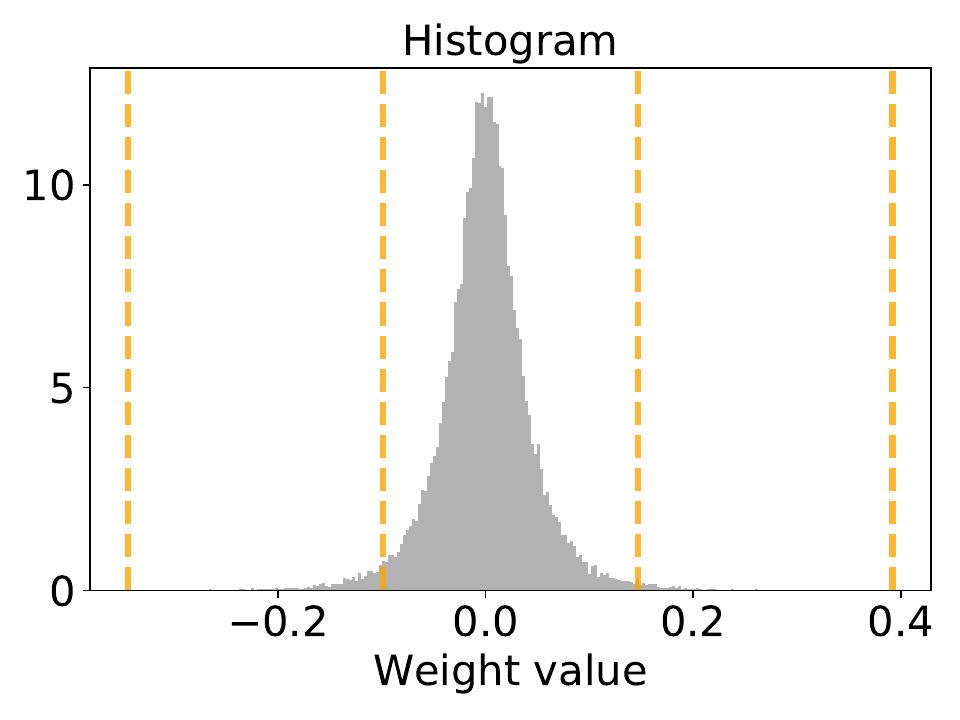}
        \caption{2-bit quantization grid\label{fig:method-weight-c}}
    \end{subfigure}
    \caption{ 
        \textbf{Layer-wise variation in error from weight quantization.}
        (a) Quantization error (QE) varies across different layers when a fixed global policy (\ie, max) is used to determine the quantization range, particularly for low bits.
        For some layers, using max does not effectively serve as a proper policy for quantization range selection.
        (b) Outliers often dominate the quantization range, leading to quantization grids being wasted on low-density areas.
        (c) At low bits, quantization grids fail to cover high-density regions adequately. Therefore, the quantization range should be adjusted for certain layers.
    }
    \label{fig:method-weight}
\end{figure}

The weight distributions of SR networks have remained relatively unexplored in previous literature.
This is because the weights in SR networks typically exhibit bell-shaped distributions, which are considered easier to quantize compared to the long-tailed, input-wise and channel-wise distinct activation distributions.
Consequently, many studies~\cite{Li2020pams,hong2022cadyq,tian2023cabm} simply adopt max quantization for weights, setting the quantization range with the maximum value of the current weight distribution.
However, we notice that this is suboptimal and may contribute to low performance when SR networks are quantized to ultra-low bits (\eg, 2-bit).
The issue arises because, in some layers, the outliers are far from the distribution mean, as visualized in \Cref{fig:method-weight-b,fig:method-weight-c}.
Thus, when the maximum value is used to determine the quantization range for such distributions, the quantization range is dominated by outliers, with quantization grids not allocated to high-density regions (\eg, near 0).
This leads to substantial quantization errors that can accumulate and degrade restoration performance.
In the case of 4-bit quantization (\Cref{fig:method-weight-b}), although grids still cover high-density regions to an extent, a number of grids are wasted on low-density areas.
The problem intensifies with low-bit (2-bit) quantization (\Cref{fig:method-weight-c}), where quantization errors become significantly larger.
This observation underscores the need for careful selection of the quantization range for layer-wise weights, particularly in low-bit quantization scenarios.

\begin{algorithm}[t]
\caption{Quantization-aware training process of ODM} 
\label{algo-odm}
\textbf{Input:} Pre-trained 32-bit network $\gP$. \\
\textbf{Output:} Quantized network $\gQ$.

\begin{algorithmic}
\For{$t=1, \cdots,$ \# iters}
\For{$i=1,\cdots,$ \# layers}
    \If{$t=1$}
        \State Initialize activation quantization range $[l_a, u_a]$
        \State Initialize weight quantization range $[\Minus u_w, u_w]$
    \EndIf
    \State Given quantization range, obtain $q(\mX_i; l_a, u_a)$ using \cref{eq:quant}
    \State Given $\mX_i$ and $q(\mX_i; l_a, u_a)$, obtain mismatch using \cref{eq:mismatch}
    \State {Adjust weight quantization range} $[\Minus u_w, u_w]$ using \cref{eq:weight-ours} 
    \State Given quantization range, obtain $q(\mW_i;\Minus u_w,u_w)$ using \cref{eq:quant}
    \State Replace $\mX_i, \mW_i$ in $\gP$ with $q(\mX_i), q(\mW_i)$
\EndFor
\State Calculate {mismatch regularization loss} (\cref{eq:loss-mismatch}) and reconstruction loss (\cref{eq:loss-rec})
\State Update parameters of $\gQ$ with two losses {cooperatively} using \cref{eq:loss-coop} 
\EndFor
\end{algorithmic}
\end{algorithm}

\subsection{Weight Clipping Correction}
\label{subsec:weight-clipping}
Also, we notice that the error from weight quantization varies across different layers, as shown in \Cref{fig:method-weight-a}.
For instance, employing the maximum value as the quantization range can be an effective policy for certain layers, yet this policy proves inadequate for others (\eg, the layer shown in \Cref{fig:method-weight-c}).
Given the unique tendency of each layer, applying a uniform global policy for selecting the quantization range across all layers is suboptimal.
Existing methods~\cite{Li2020pams,hong2022cadyq,tian2023cabm} utilize a fixed global policy throughout training to set the quantization range clipping parameter $u_w$ as follows:
\begin{equation}
\label{eq:weight-global}
    u_w^{t} = f(\mW^{t}),
\end{equation}
where the global policy $f(\cdot)$ is the same function for all layers (\eg, $max(\cdot)$).
A simple solution to accommodate layer-specific variations is to make the clipping parameter $u_w$ for each layer a learnable parameter~\cite{esser2019learned}:
\begin{equation}
\label{eq:weight-lsq}
    u_w^{t+1} = u_w^t - \beta^t \cdot \nabla_{u_w} \mathcal{L}_R(u_w^t),
\end{equation}
where $\beta^t$ denotes the learning rate.
This process determines the clipping parameter $u_w^{t}$ to quantize $\mW^t$ based on the weight of the previous iteration, $\mW^{t-1}$.
However, since the weight is also updated at iteration step $t$ ($\mW^{t-1}\shortrightarrow \mW^{t}$), a mismatch occurs between the current weight and the weight quantization range derived from the previous weight.
To address this, we first obtain the quantization range by applying the global policy to the current weight, then adjust the range with a learnable parameter that accounts for layer-specific tendencies.
Our clipping parameter is formulated as follows:
\begin{equation}
\label{eq:weight-ours}
    u_w^{t+1} = f(\mW^{t+1}) \cdot (\gamma_w^t - \beta^t \cdot \nabla_{\gamma_w} \mathcal{L}_R(\gamma_w^t) ),
\end{equation}
where $\gamma_w$ is the learnable parameter for each layer representing the layer-wise adjustment.
Each $\gamma_w$ is initially set to 1.
To prevent outliers from dominating the initial quantization range, we set the global policy $f(\cdot)$ as the $j$-th percentile function.
Our clipping correction scheme introduces only one additional parameter per layer; thus, the overall computational overhead is minimal. For further details, please refer to \Cref{subsec:complexity}.

\section{Experiments}
\label{sec:experiments}

\begin{table}[t]
\setlength{\tabcolsep}{1.2mm}
\renewcommand\TPTminimum{\linewidth}
\caption{ \textbf{Quantitative comparisons on EDSR} of scale 4
}
\label{tab:exp-edsr}
\begin{threeparttable}
\centering
    \makebox[\linewidth]{\scriptsize
    \scalebox{1.1}{
    \begin{tabular}{l c cc cc cc cc}
        \toprule
        \multirow{2}{*}{Model} & \multirow{2}{*}{Bit} & 
        \multicolumn{2}{c}{Set5} & \multicolumn{2}{c}{Set14} & \multicolumn{2}{c}{B100} & \multicolumn{2}{c}{Urban100} \\
        \cmidrule(lr){3-4} \cmidrule(lr){5-6} \cmidrule(lr){7-8} \cmidrule(lr){9-10}& 
         & PSNR & SSIM & PSNR & SSIM & PSNR & SSIM & PSNR & SSIM \\
        \midrule
        EDSR   &32 &32.10&0.894 &28.58&0.781 &27.56&0.736 &26.04&0.785\\
        \midrule
        EDSR-PAMS   &4  &31.59&0.885 &28.20&0.773 &27.32&0.728 &25.32&0.762\\
        EDSR-DAQ &4  &31.85&0.887 &28.38&0.776 &27.42&0.732 &25.73&0.772\\
        EDSR-DDTB&4  &31.85&0.889 &28.39&0.777 &27.44&0.732 &25.69&0.774\\
        EDSR-ODM (Ours) &4 &\textbf{32.00}& \textbf{0.891} &\textbf{28.47}& \textbf{0.779}& \textbf{27.51} & \textbf{0.735} & \textbf{25.80} & \textbf{0.778}\\
        \midrule
        EDSR-PAMS   &3 &27.25&0.780 &25.24&0.673 &25.38&0.644 &22.76&0.641\\
        EDSR-DAQ   &3 &31.66&0.884 &28.19&0.771 &27.28&0.728 &25.40&0.762\\
        EDSR-DDTB&3 &31.52&0.883 &28.18&0.771 &27.30&0.727 &25.33&0.761\\
        EDSR-ODM (Ours) &3 &\textbf{31.85}& \textbf{0.888} & \textbf{28.38} & \textbf{0.776} & \textbf{27.43} & \textbf{0.732} & \textbf{25.59} & \textbf{0.771} \\
        \midrule
        EDSR-PAMS   &2 &29.51&0.835 &26.79&0.734 &26.45&0.696 &23.72&0.688\\
        EDSR-DAQ $^\dagger$&2 &31.01&0.871 &27.89&0.762 &27.09&0.719 &24.88&0.740\\
        EDSR-DDTB&2 &30.97&0.876 &27.87&0.764 &27.09&0.719 &24.82&0.742\\
        EDSR-ODM (Ours) &2 &\textbf{31.50}& \textbf{0.882} & \textbf{28.14} & \textbf{0.770} & \textbf{27.27} & \textbf{0.726} & \textbf{25.17} & \textbf{0.755} \\
        \bottomrule
    \end{tabular}
    }
    }
\begin{tablenotes}
    \item[$\dagger$]\scriptsize{ We note that the reported results of DAQ~\cite{hong2022daq} are obtained using EDSR of 32 residual blocks. For a fair comparison, we reproduce EDSR-DAQ using EDSR of 16 residual blocks (\ie, EDSR-baseline).
    } 
\end{tablenotes}
\end{threeparttable}
\end{table}

\begin{table}[t]
\setlength{\tabcolsep}{1.2mm}
\caption{ \textbf{Quantitative comparisons on RDN} of scale 4
}
\centering
\makebox[\linewidth]{\scriptsize
    \scalebox{1.1}{
    \begin{tabular}{l c cc cc cc cc}
        \toprule
        \multirow{2}{*}{Model} & \multirow{2}{*}{Bit} & 
        \multicolumn{2}{c}{Set5} & \multicolumn{2}{c}{Set14} & \multicolumn{2}{c}{B100} & \multicolumn{2}{c}{Urban100} \\
        \cmidrule(lr){3-4} \cmidrule(lr){5-6} \cmidrule(lr){7-8} \cmidrule(lr){9-10}& 
         & PSNR & SSIM & PSNR & SSIM & PSNR & SSIM & PSNR & SSIM \\
        \midrule 
        RDN   &32 &32.24&0.896 &28.67&0.784 &27.63&0.738 &26.29&0.792\\
        \midrule
        RDN-PAMS   &4  &30.44&0.862 &27.54&0.753 &26.87&0.710 &24.52&0.726\\
        RDN-DAQ   &4  &31.91&0.889 &28.38&0.775 &27.38&0.733 &25.81&0.779\\
        RDN-DDTB&4  &31.97&0.891 &28.49&0.780 &27.49&0.735 &25.90&0.783\\
        RDN-ODM (Ours) &4 &\textbf{32.06}&\textbf{0.892} &\textbf{28.49}&\textbf{0.780} &\textbf{27.52} &\textbf{0.736} &\textbf{25.88} &\textbf{0.783}\\
        \midrule
        RDN-PAMS &3 &29.54&0.838 &26.82&0.734 &26.47&0.696 &23.83&0.692\\
        RDN-DAQ  &3 &31.57&0.883 &28.18&0.771 &27.27&0.728 &25.47&0.765\\
        RDN-DDTB&3 &31.49&0.883 &28.17&0.772 &27.30&0.728 &25.35&0.764\\
        RDN-ODM (Ours) &3 &\textbf{31.79}&\textbf{0.887} &\textbf{28.33} &\textbf{0.776} &\textbf{27.42} & \textbf{0.732} &\textbf{25.51} & \textbf{0.770}\\
        \midrule
        RDN-PAMS   &2 &29.73&0.843 &26.96&0.739 &26.57&0.700 &23.87&0.696\\
        RDN-DAQ   &2 &30.71&0.866 &27.61&0.755 &26.93&0.715 &24.71&0.731\\
        RDN-DDTB&2 &30.57&0.867 &27.56&0.757 &26.91&0.714 &24.50&0.728\\
        RDN-ODM (Ours) &2 &\textbf{31.37}& \textbf{0.880}& \textbf{28.08} & \textbf{0.770}& \textbf{27.24} & \textbf{0.727} &\textbf{25.09} & \textbf{0.755}\\
        \bottomrule
    \end{tabular}
}
}
\label{tab:exp-rdn}
\end{table}

\begin{table}[t]
\centering
\setlength{\tabcolsep}{1.2mm}
\caption{ \textbf{Quantitative comparisons on SwinIR} of scale 4
}
\makebox[\linewidth]{\scriptsize
    \scalebox{1.1}{
    \begin{tabular}{l c cc cc cc cc}
        \toprule
        \multirow{2}{*}{Model} & \multirow{2}{*}{Bit} & 
        \multicolumn{2}{c}{Set5} & \multicolumn{2}{c}{Set14} & \multicolumn{2}{c}{B100} & \multicolumn{2}{c}{Urban100} \\
        \cmidrule(lr){3-4} \cmidrule(lr){5-6} \cmidrule(lr){7-8} \cmidrule(lr){9-10}& 
         & PSNR & SSIM & PSNR & SSIM & PSNR & SSIM & PSNR & SSIM \\
        \midrule
        SwinIR   &32 &32.44&0.898 &28.77&0.786 &27.69&0.741 &26.47&0.798\\
        \midrule
        SwinIR-PAMS &4  &31.99&0.890 &28.50&0.779 &27.49&0.735 &25.86&0.780\\
        SwinIR-DAQ  &4  &31.82&0.887 &28.34&0.775 &27.37&0.730 &25.68&0.772\\
        SwinIR-DDTB &4  &32.09&0.891 &28.55&0.780 &27.54&0.735 &26.01&0.783\\
        SwinIR-ODM (Ours) & 4 &\textbf{32.17}& \textbf{0.892} &\textbf{28.59}&  \textbf{0.781}& \textbf{27.56} & \textbf{0.736} & \textbf{26.06} & \textbf{0.785}\\
        \midrule
        SwinIR-PAMS &3 &31.62&0.884 &28.23&0.771 &27.31&0.728 &25.38&0.762\\
        SwinIR-DAQ  &3 &31.50&0.882 &27.99&0.770 &27.12&0.727 &25.29&0.761\\
        SwinIR-DDTB &3 &31.80&0.887 &28.34&0.775 &27.40&0.731 &25.63&0.771\\
        SwinIR-ODM (Ours) & 3 &\textbf{31.94}& \textbf{0.889} & \textbf{28.39} & \textbf{0.777} & \textbf{27.45} & \textbf{0.733} & \textbf{25.72} & \textbf{0.775} \\
        \midrule
        SwinIR-PAMS &2 &29.48&0.834 &26.79&0.733 &26.46&0.698 &23.72&0.688\\
        SwinIR-DAQ  &2 &29.10&0.824 &26.55&0.725 &26.30&0.691 &23.51&0.678\\
        SwinIR-DDTB &2 &31.01&0.873 &27.80&0.762 &27.04&0.719 &24.79&0.739\\
        SwinIR-ODM (Ours) & 2 &\textbf{31.44}& \textbf{0.880} & \textbf{28.06} & \textbf{0.769} & \textbf{27.23} & \textbf{0.725} & \textbf{25.14} & \textbf{0.754} \\
        \bottomrule
    \end{tabular}
}
}
\label{tab:exp-swinir}
\end{table}

The efficacy and adaptability of the proposed quantization framework, ODM, are assessed through its application across several SR networks.
The experimental settings are described (\Cref{subsec:setting}), and quantitative (\Cref{subsec:quan}), qualitative (\Cref{subsec:qual}), and complexity (\Cref{subsec:complexity}) evaluations are conducted on various SR networks. 
Ablation studies are conducted to examine each component of the framework (\Cref{subsec:ablation}).

\subsection{Implementation Details}
\label{subsec:setting}

\subsubsection{Models and Training.}
The proposed framework is applied directly to existing representative SR networks that produce satisfactory SR results, but involve heavy computations: EDSR (baseline)~\cite{lim2017enhanced} and RDN~\cite{zhang2018residual}.
Furthermore, we apply our method to the Transformer-based SR model, SwinIR-S~\cite{liang2021swinir}.
Following prior works on SR quantization~\cite{Li2020pams,ma2019efficient,xin2020binarized,hong2022daq,zhong2022ddtb,hong2022cadyq}, weights and activations of the high-level feature extraction module are quantized, which is the most computationally demanding. 
Training and validation are conducted using the DIV2K~\cite{agustsson2017ntire} dataset. 
ODM trains the network for 60K iterations with a batch size of 8.
The weights are updated with an initial learning rate of $\beta^0\Equal10^{-4}$. For cooperative update, we update clipping parameters with $10\cdot\beta^0$ initial learning rate.
The learning rates are halved every 15K iteration.
The hyperparameter for the percentile is set to $j=99$, and to balance the gradient of the loss terms, we set $\lambda_{R}\Equal1$ and $\lambda_{M}\Equal10^{-5}$.
Specially, we set $\lambda_{M}\Equal10^{-6}$ for RDN whose overall mismatch is large.
Ablation studies on the hyperparameters are in the supplementary material.
All our experiments are implemented using PyTorch and run on an RTX 2080Ti GPU.

\subsubsection{Evaluation.}
We evaluate our framework on the standard benchmark (Set5~\cite{bevilacqua2012low}, Set14~\cite{ledig2017photo}, BSD100~\cite{martin2001database}, and Urban100~\cite{huang2015single}) by measuring the peak signal-to-noise ratio (PSNR) and the structural similarity index (SSIM~\cite{wang2004image}).
To assess the computational complexity of our framework, we measure bitOPs and storage size. 
BitOPs refers to the number of operations weighted by the bit-widths of the two operands.
Storage size is calculated as the number of stored parameters weighted by the precision of each parameter value.

\subsection{Quantitative Results}
\label{subsec:quan}

To evaluate the effectiveness of our proposed scheme, we compare the results with existing SR quantization works using their official codes: PAMS~\cite{Li2020pams}, DAQ~\cite{hong2022daq}, and DDTB~\cite{zhong2022ddtb}. 
For a fair comparison, we reproduce other methods using the same training iterations, 60K iterations.
The supplementary materials provide additional experiments that further demonstrate the applicability of ODM, including results on 300K iterations, scale 2, and fully quantized settings.

\noindent \textbf{EDSR. }
As shown in \Cref{tab:exp-edsr}, ODM outperforms other methods in the 4, 3, and 2-bit settings, and notably, the improvement is significant for 2-bit, achieving a gain of more than 0.49 dB for Set5.
We notice that 4-bit EDSR-ODM achieves closer accuracy to 32-bit EDSR, with a marginal difference of 0.1 dB for Set5.
This indicates that ODM can effectively bridge the gap between the quantized network and the 32-bit network.

\noindent \textbf{RDN. }
Similarly, \Cref{tab:exp-rdn} compares the results on RDN, whose computational complexity is more burdensome than EDSR.
The results show that ODM consistently achieves superior performance on 4, 3, and 2-bit quantization.
The gain over existing methods is especially large for the 2-bit setting, where it exceeds 0.66 dB for Set5.

\noindent \textbf{SwinIR. }
Furthermore, we evaluate our framework on the Transformer-based architecture, SwinIR.
The linear and convolutional layers of SwinIR are quantized.
According to \Cref{tab:exp-swinir}, ODM is also proven effective in quantizing SwinIR across all bit settings, where the improvement is most notable in the 2-bit setting (0.43 dB).

\noindent \textbf{Comparison with QuantSR. }
We also compare our method with the concurrent work, QuantSR~\cite{qin2024quantsr}.
As the training code of QuantSR has not been released, we base our comparison on the reported performance.
For a fair comparison with QuantSR's reported performance, we also train our model for 300K iterations on SRResNet~\cite{ledig2017photo} and SwinIR~\cite{liang2021swinir}.
In \Cref{tab:exp-quantsr}, the results demonstrate that our method achieves better results than QuantSR;
compared to QuantSR, ours shows a gain of 0.51 dB on 2-bit SRResNet and a gain of 0.14 dB on 2-bit SwinIR for Set5.

\begin{table}[t]
\setlength{\tabcolsep}{1.2mm}
\caption{ \textbf{Quantitative comparisons with QuantSR} on SRResNet and SwinIR of scale 4. For a fair comparison, our model (ODM$^*$) is trained for 300K iterations following QuantSR.
}
\centering
\makebox[\linewidth]{\scriptsize
    \scalebox{1.1}{
    \begin{tabular}{l c cc cc cc cc}
        \toprule
        \multirow{2}{*}{Model} & \multirow{2}{*}{Bit} & 
        \multicolumn{2}{c}{Set5} & \multicolumn{2}{c}{Set14} & \multicolumn{2}{c}{B100} & \multicolumn{2}{c}{Urban100} \\
        \cmidrule(lr){3-4} \cmidrule(lr){5-6} \cmidrule(lr){7-8} \cmidrule(lr){9-10}& 
         & PSNR & SSIM & PSNR & SSIM & PSNR & SSIM & PSNR & SSIM \\
        \midrule
        SRResNet   &32 &32.07&0.893 &28.50&0.780 &27.52&0.735 &25.86&0.779\\
        \midrule
        SRResNet-QuantSR & 2 & 31.30 & 0.882 & 28.08 & 0.769 & 27.23 & 0.725 & 25.13 & 0.754\\
        SRResNet-ODM$^*$(Ours) & 2 &\textbf{31.81}&\textbf{0.888} &\textbf{28.32}&\textbf{0.774} &\textbf{27.38}&\textbf{0.730} &\textbf{25.54}&\textbf{0.767}\\
        \midrule
        SwinIR   &32 & 32.44 & 0.898 & 28.77 & 0.786 & 27.69 & 0.741 & 26.47 & 0.798\\
        \midrule
        SwinIR-QuantSR & 2 & 31.53 & 0.885 & 28.16 & 0.772 & 27.28 & 0.727 & 25.26 & 0.761\\
        SwinIR-ODM$^*$(Ours) & 2 &\textbf{31.67}& \textbf{0.885} & \textbf{28.23} & \textbf{0.772} & \textbf{27.33} & \textbf{0.728} & \textbf{25.36} & \textbf{0.762} \\
        \bottomrule
    \end{tabular}
}
}
\label{tab:exp-quantsr}
\end{table}

\newcommand{\co}{black}
\begin{table}[t]
    \centering
    \setlength{\tabcolsep}{1.4mm}
    \renewcommand{\arraystretch}{1.2}
    \aboverulesep=0ex
    \belowrulesep=0ex
    \caption{ \textbf{Computational complexity comparison} with SR quantization methods
    }
    \subfloat[{EDSR}]{
        \resizebox{0.5\textwidth}{!}{
            \begin{tabular}{l|c|cc|cc}
                \toprule
                Model & Bit & Storage size & BitOPs & PSNR & SSIM \\
                \midrule
                EDSR & 32 & 1517.6K & 527.1T & 32.10 & 0.894 \\
                \midrule
                EDSR-PAMS & 2 & {411.7K} & {{215.1T}} & 29.51 & 0.835\\
                EDSR-DAQ & 2 & 411.7K & \underline{215.6T}& 31.01 & 0.871 \\
                EDSR-DDTB & 2 & \underline{413.6K} & 215.1T& 30.97 & 0.876 \\
                EDSR-ODM (Ours) & 2 & {\textbf{411.8K}} & {\textbf{215.1T}} & {\textbf{31.50}} & {\textbf{0.882}} \\
                \bottomrule
            \end{tabular}
        }
    }
    \subfloat[RDN]{
        \resizebox{0.5\textwidth}{!}{
            \begin{tabular}{l|c|cc|cc}
                \toprule
                Model & Bit & Storage size & BitOPs & PSNR & SSIM \\
                \midrule
                RDN & 32 &  22271.1K &  6032.9T & 32.24 & 0.896 \\
                \midrule
                RDN-PAMS & 2 & 1715.9K & 236.6T & 29.73 & 0.843\\
                RDN-DAQ  & 2 & 1715.9K & \underline{287.7T} & 30.71 & 0.866 \\
                RDN-DDTB & 2 & \underline{1761.6K} & 236.6T & 30.57 & 0.867 \\
                RDN-ODM (Ours) & 2 & \textbf{1716.1K} & \textbf{236.6T} & \textbf{31.37} & \textbf{0.880} \\
                \bottomrule
            \end{tabular}
        }
    }
    \label{tab:exp-complexity}
\end{table}
\begin{figure}[t]
\centering
\begin{center}
    \setlength{\tabcolsep}{0.1cm}
    \newcommand{\w}{0.18\linewidth}
    \begin{tabular}{ccccc}
        \centering
        \includegraphics[width=\w]{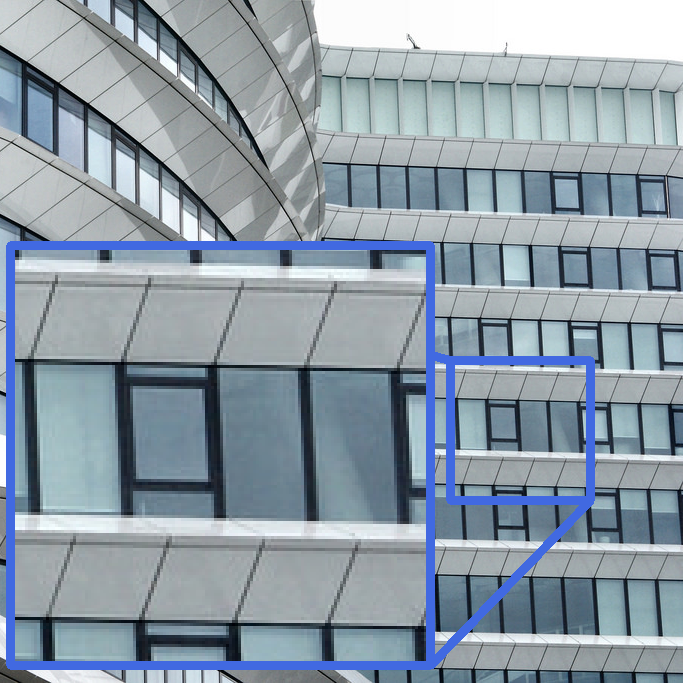} &
        \includegraphics[width=\w]{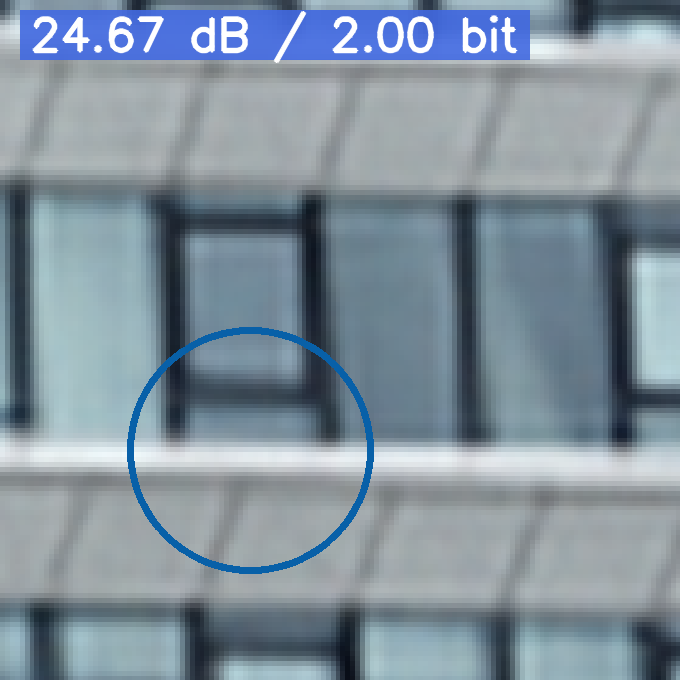} &
        \includegraphics[width=\w]{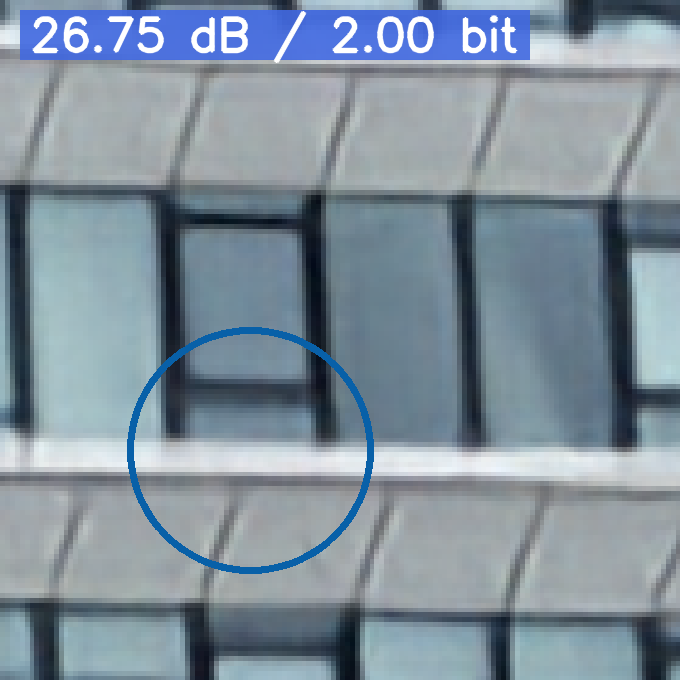} &
        \includegraphics[width=\w]{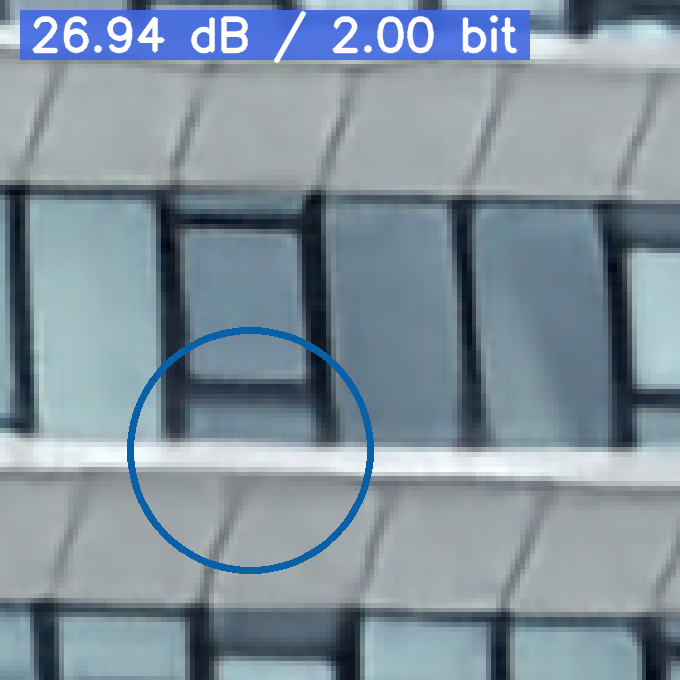} &
        \includegraphics[width=\w]{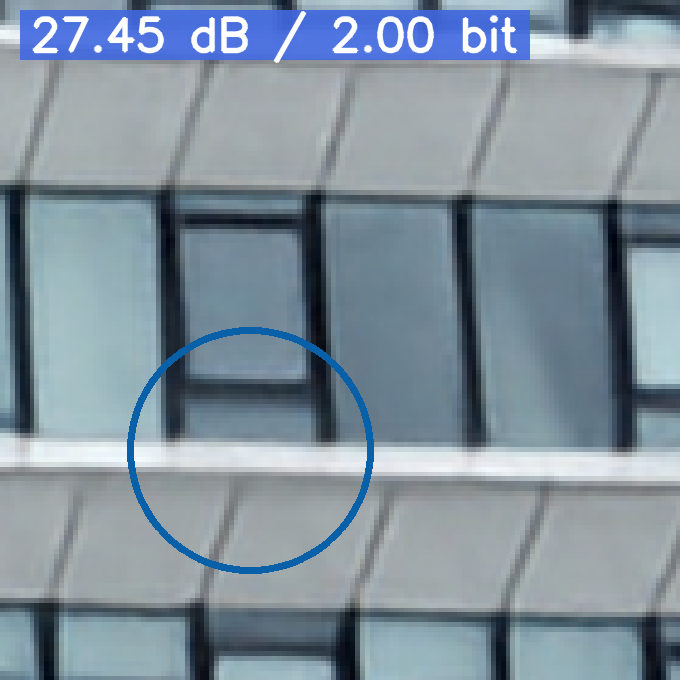} \\
        \scriptsize GT (img052) & \scriptsize EDSR-PAMS & \scriptsize EDSR-DAQ &\scriptsize EDSR-DDTB &\scriptsize EDSR-ODM (Ours) \\
        \includegraphics[width=\w,height=\w]{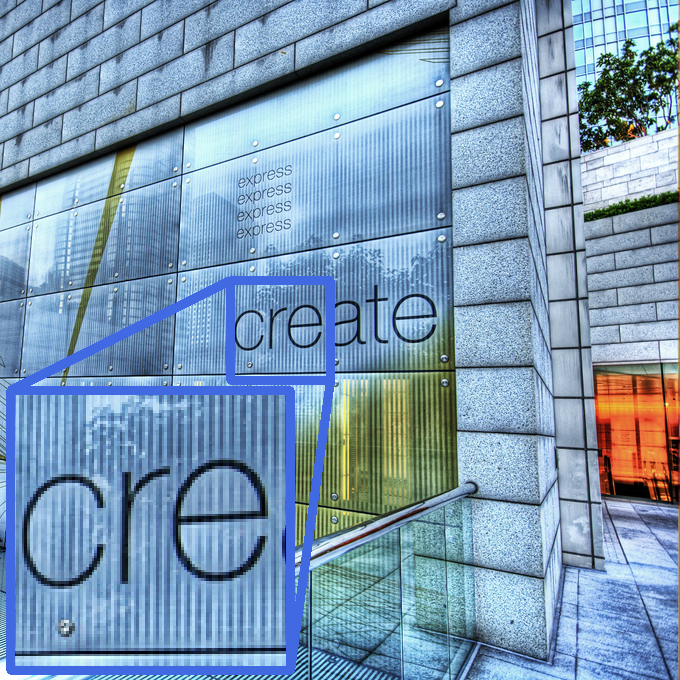} & 
        \includegraphics[width=\w,height=\w]{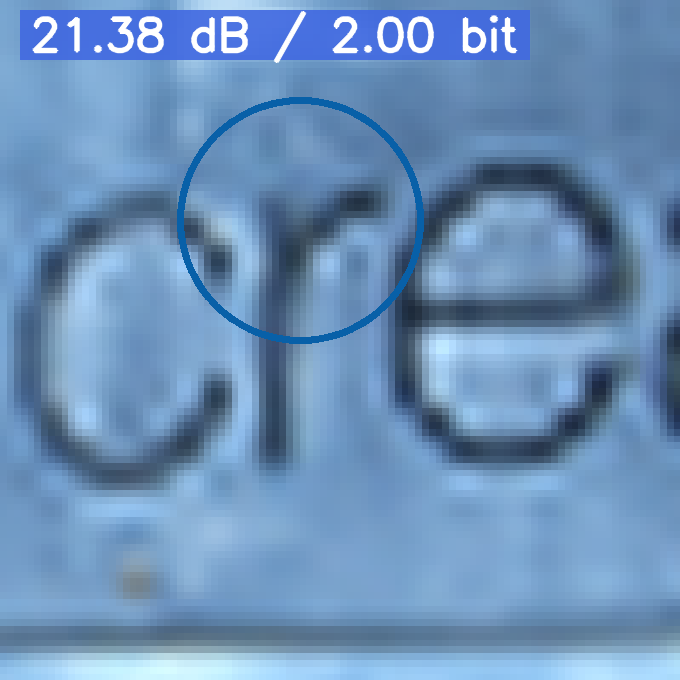} & 
        \includegraphics[width=\w,height=\w]{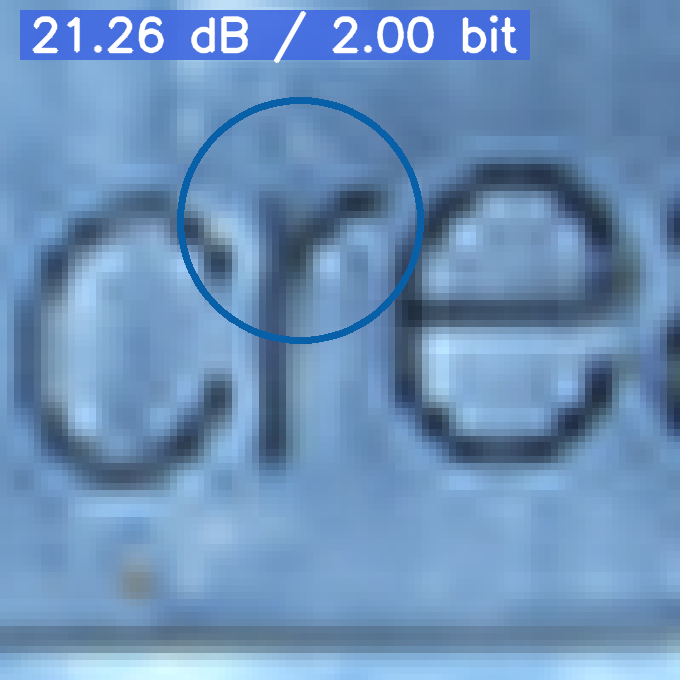} & 
        \includegraphics[width=\w,height=\w]{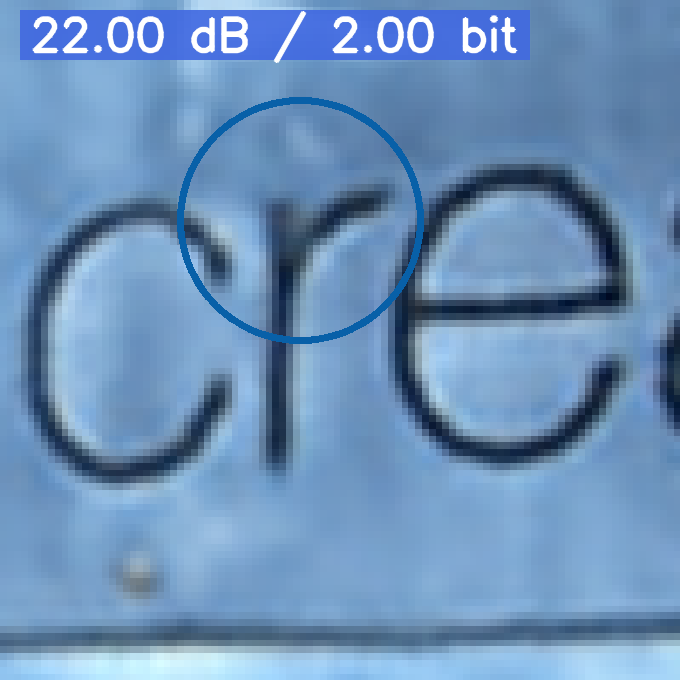} & 
        \includegraphics[width=\w,height=\w]{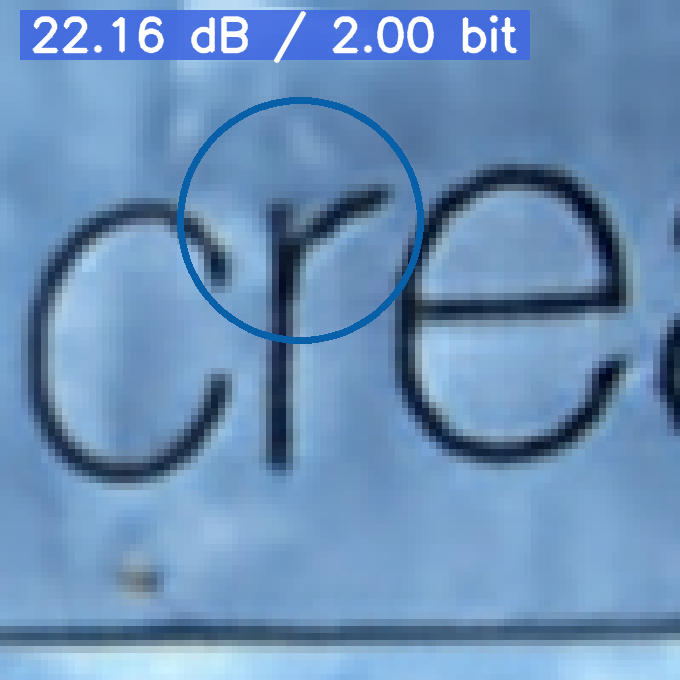} \\
        \scriptsize GT (img060) & \scriptsize SwinIR-PAMS & \scriptsize SwinIR-DAQ &\scriptsize SwinIR-DDTB &\scriptsize SwinIR-ODM (Ours) \\
        \includegraphics[width=\w]{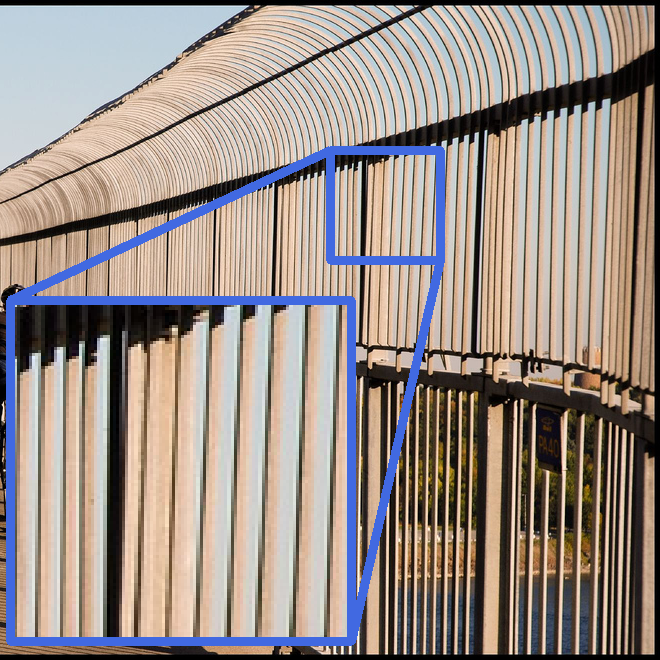} &
        \includegraphics[width=\w]{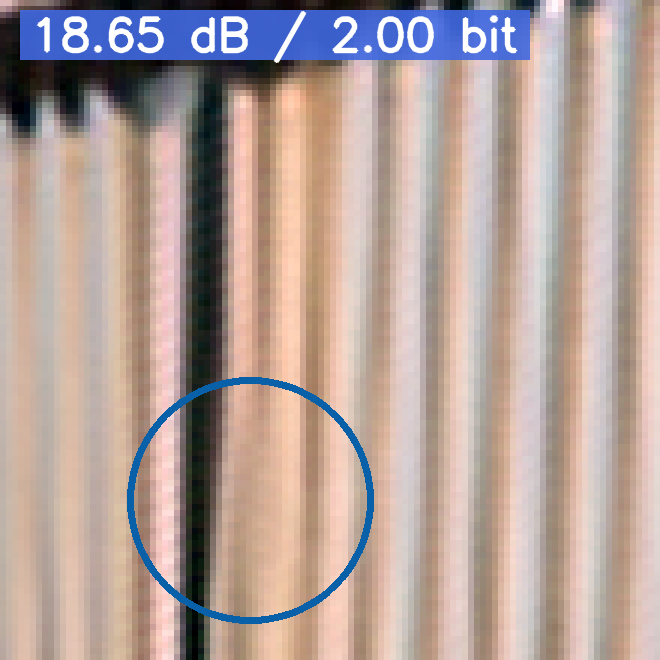} &
        \includegraphics[width=\w]{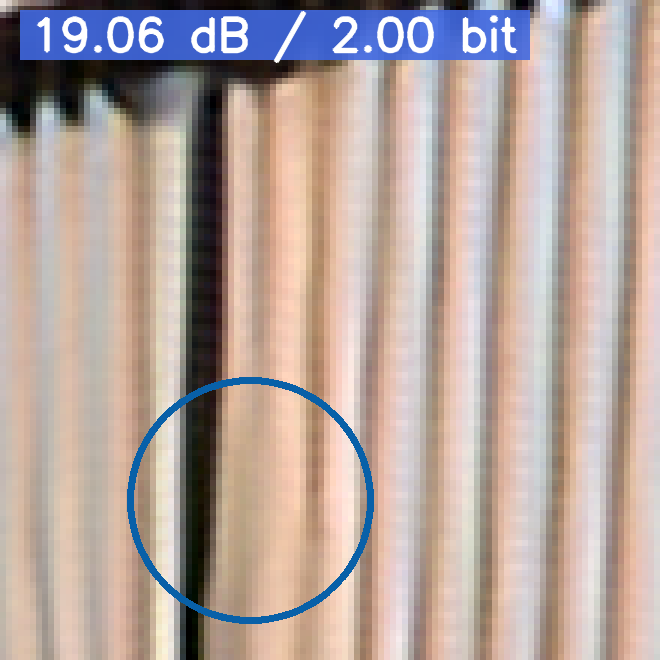} &
        \includegraphics[width=\w]{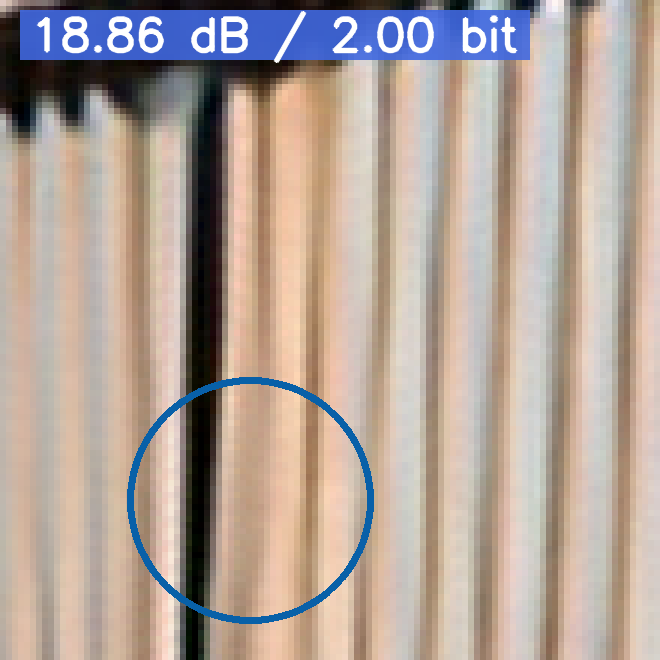} &
        \includegraphics[width=\w]{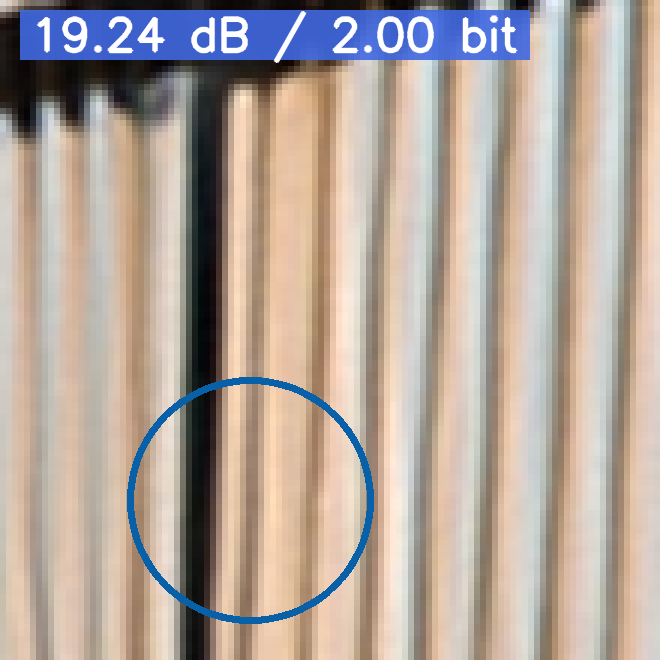} \\
        \scriptsize GT (img024) & \scriptsize RDN-PAMS &\scriptsize RDN-DAQ &\scriptsize RDN-DDTB &\scriptsize RDN-ODM (Ours) \\
    \end{tabular}
    \caption{
    \textbf{Qualitative results} on Urban100 with EDSR, SwinIR, and RDN-based models
    }
    \label{fig:exp-qualitative}
\end{center}
\end{figure}

\subsection{Complexity Analysis}
\label{subsec:complexity}
Along with the accuracy of SR, we also evaluate the computational complexity of our framework in \Cref{tab:exp-complexity}.
We measure the storage size for the model weights and the bitOPs required for generating a 1920$\times$1080 image with $\times4$ SR network.
Overall, our framework, ODM, achieves higher restoration accuracy with similar or fewer computational resources.
Specifically, our weight-clipping correction involves an additional storage size of 0.06K / 0.15K for EDSR / RDN.
Since the correction process can be predetermined before test time, no extra bitOPs are required.
Compared to existing methods, our method achieves $\times$31.7 smaller storage size overhead than DDTB and 0.5T fewer bitOPs than DAQ on EDSR.
For RDN, the gap is larger; our method's overhead is $\times$304.7 smaller in storage size than DDTB and 51.1T fewer in bitOPs than DAQ.
Moreover, we note that DAQ adopts a channel-wise dynamic quantization function and DDTB an asymmetric weight quantization function, which are not favorably supported by hardware.
Despite incurring a minor additional storage size of 0.06K / 0.15K over PAMS, the accuracy improvement over PAMS is significant (+1.99 dB / +1.64 dB).
The results prove that our method achieves significant accuracy gain with minimal or no computational overhead.

\subsection{Qualitative Results}
\label{subsec:qual}

\Cref{fig:exp-qualitative} provides qualitative results and comparisons with the output images of quantized EDSR, RDN, and SwinIR.
Our method, ODM, produces visually cleaner output images compared to existing methods.
In contrast, existing methods, especially PAMS, suffer from blurred lines or artifacts.
These qualitative results stress the importance of alleviating the distribution mismatch problem in SR networks.

\subsection{Ablation Study}
\label{subsec:ablation}

In \Cref{tab:exp-ablation}, we verify the importance of each attribute of our framework: cooperative mismatch regularization and weight clipping correction.
According to the results, each attribute individually improves baseline accuracy.
Weight clipping correction improves the baseline by {+1.18 dB} for Set5, indicating that considering both the layer-wise trend and the current weight distribution is important for weight quantization.
Also, while directly integrating mismatch regularization with the reconstruction loss (Model (\textbf{d})) rather degrades performance by -0.13 dB, our cooperative scheme (Model (\textbf{e})) improves the SR accuracy by {+0.51 dB}.
This shows that reducing the mismatch in both activations and weights is important for accurately quantizing SR networks.

\definecolor{brickred}{rgb}{0.8, 0.25, 0.33}
\definecolor{brickred2}{rgb}{0.25, 0.8, 0.33}
\begin{table}[!ht]
    \renewcommand{\arraystretch}{1.1}
    \setlength{\tabcolsep}{1.2mm}
    \centering
    \aboverulesep=0ex
    \belowrulesep=0ex
    \caption{ \textbf{Ablation study on each attribute of our framework.} 
    \textit{WCC} refers to weight clipping correction and \textit{MR} refers to mismatch regularization.
    \textit{Cooperative} denotes whether the mismatch regularization is cooperatively used with the reconstruction loss.
    Non-\textit{cooperative} \textit{MR} denotes that the two losses are simply used together.
    }
    \scalebox{0.85}{
    \begin{tabular}{@{}c | ccc | cc}
        \toprule
        {Model} & {WCC} & {Cooperative} & {MR} & {Set5 (PSNR / SSIM)} & {Urban100 (PSNR / SSIM)}\\
        \midrule
        \textbf{(a)} & - & - & -       & 29.94 / 0.848 & 23.99 / 0.703 \\
        \textbf{(b)} & - & \cm & \cm   & 30.34 / 0.859 & 24.27 / 0.715 \\
        \textbf{(c)} & \cm & - & -     & 31.12 / 0.876 & 24.91 / 0.746 \\
        \textbf{(d)} & \cm & - & \cm   & 30.99 / 0.871 & 24.79 / 0.735 \\
        \textbf{(e)} & \cm & \cm & \cm & \textbf{31.50 / 0.882} & \textbf{25.17 / 0.755} \\
        \bottomrule
    \end{tabular}
    }
    \label{tab:exp-ablation}
\end{table}

\begin{figure}[!ht]
\centering
\begin{center}
    \setlength{\tabcolsep}{0.05cm}
    \newcommand{\w}{0.22\linewidth}
    \begin{tabular}{cccc}
        \centering
        \includegraphics[width=\w]{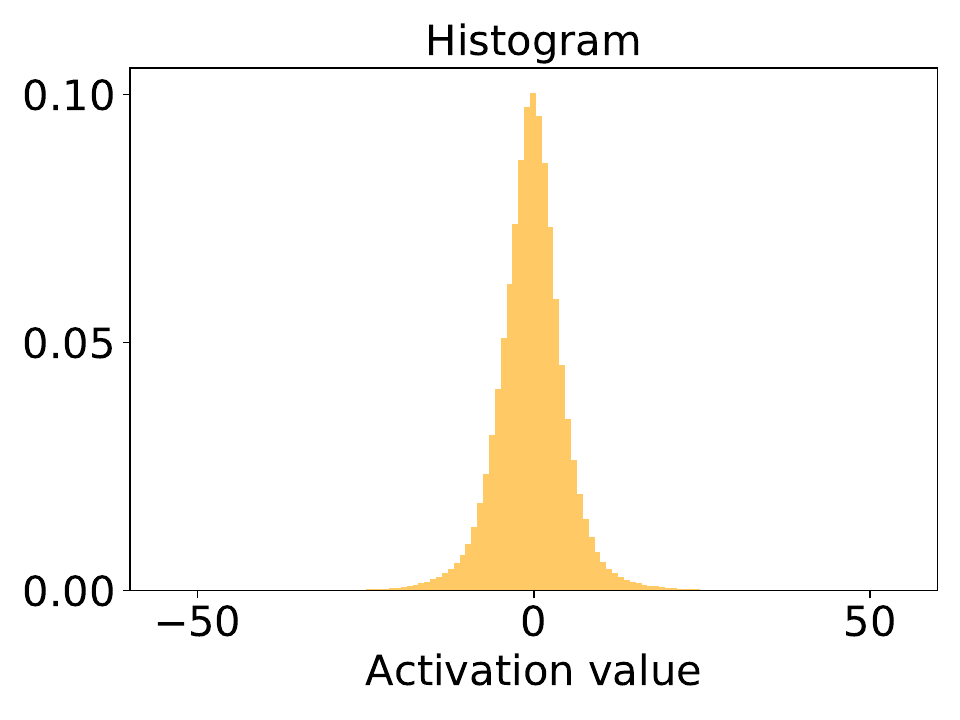} &
        \includegraphics[width=\w]{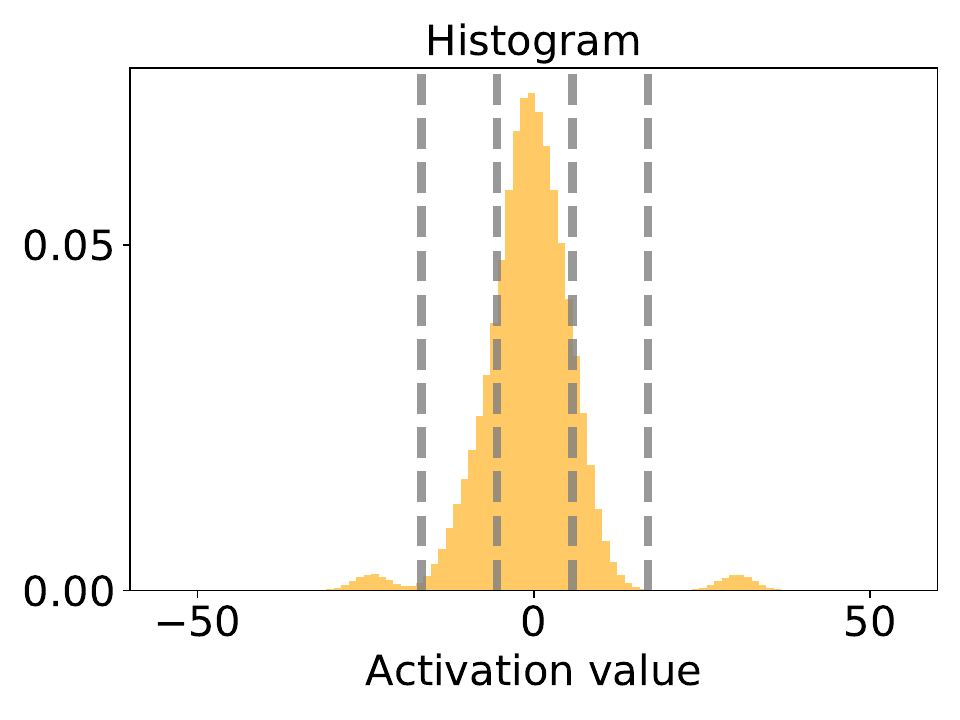} &
        \includegraphics[width=\w]{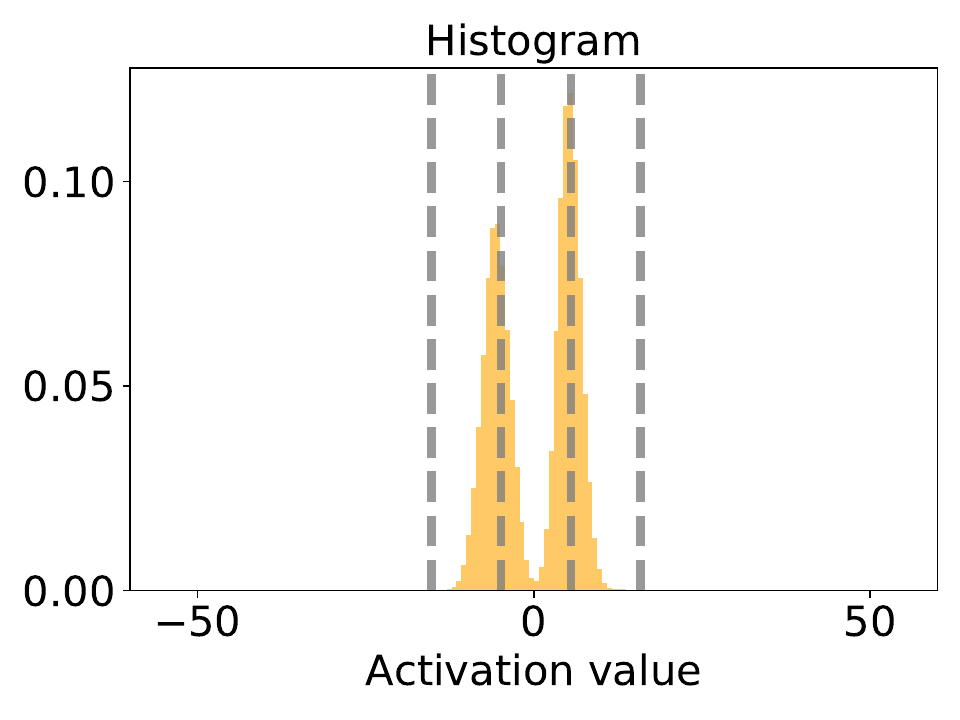} &
        \includegraphics[width=\w]{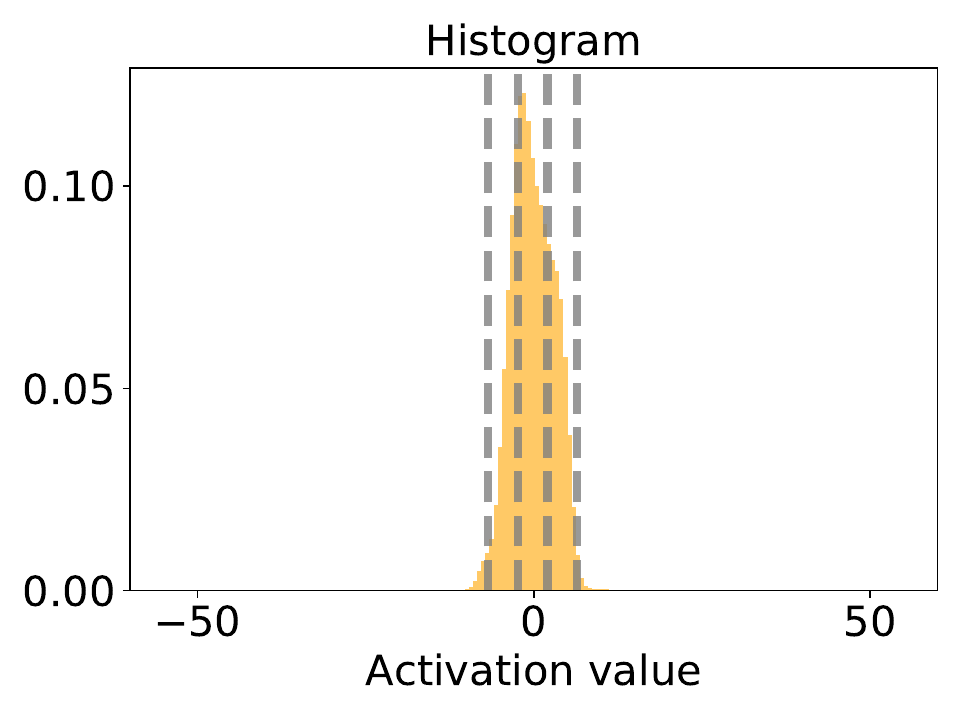} \\
        \scriptsize{32-bit network} & \scriptsize{$\mathcal{L}_{R}$} & \scriptsize{$\mathcal{L}_{R}+\mathcal{L}_{M}$} & \scriptsize{Cooperative MR} \\
    \end{tabular}
    \caption{
    \textbf{Distribution of activations before quantization.}
    Using our cooperative mismatch regularization results in distributions more robust to low-bit quantization.
    8th conv layer of EDSR-ODM (2-bit) on `baby' (Set5) are visualized.
    }
    \label{fig:exp-dist}
\end{center}
\end{figure}

Furthermore, we visualize the feature distributions in \Cref{fig:exp-dist} to validate the importance of our cooperative regularization term.
After training only with the reconstruction loss, the outliers remain far from the quantization grid.
When mismatch regularization loss is simply added to the original reconstruction loss, the activation distribution falls into a narrower range and resembles a multi-modal distribution near the quantization grids. 
Although this distribution is quantization-friendly, it significantly deviates from the original activation of the 32-bit network and removes originally dense values (near 0).
This can lead to an accuracy drop, as demonstrated in \Cref{tab:exp-ablation} (\textbf{d}).
In contrast, our cooperative mismatch regularization results in quantization-friendly distribution while largely preserving the activations in the dense region of the original 32-bit network.

\section{Conclusion}
\label{sec:conclusion}
SR networks suffer accuracy loss from quantization due to the inherent mismatch in feature distributions.
Instead of employing resource-demanding dynamic modules to handle distinct distributions during test time, we introduce a new quantization-aware training technique that alleviates this mismatch through distribution optimization.
We utilize cooperative mismatch regularization to update the SR network to be quantization-friendly and accurate.
Additionally, to address the mismatch in layer-wise weights, we propose a weight-clipping correction strategy.
These straightforward solutions effectively reduce the distribution mismatch with minimal computational overhead.

\section*{Acknowledgment}
\label{sec:acknowledgment}
This work was supported in part by the IITP grants [No. 2021-0-01343, Artificial Intelligence Graduate School Program (Seoul National University), No.2021-0-02068, and No.2023-0-00156], the NRF grant [No.2021M3A9E4080782] funded by the Korean government (MSIT).

%
%
\bibliographystyle{splncs04}
\bibliography{main}

\clearpage
\title{Supplementary Material for\texorpdfstring{\\}{} ``Overcoming Distribution Mismatch\texorpdfstring{\\}{} in Quantizing Image Super-Resolution Networks''}
\titlerunning{ODM}
\author{Cheeun Hong\inst{1}\orcidlink{0009-0009-3480-748X} \and
Kyoung Mu Lee\inst{1,2}\orcidlink{0000-0001-7210-1036}}
\authorrunning{C.~Hong and K.M.~Lee}
\institute{
Dept. of ECE \& ASRI, \email{\{cheeun914, kyoungmu\}@snu.ac.kr} \and
IPAI, Seoul National University
}
\maketitle

\renewcommand{\thetable}{S\arabic{table}}
\renewcommand{\thesection}{S\arabic{section}}
\renewcommand{\thefigure}{S\arabic{figure}}
\renewcommand{\theequation}{S\arabic{equation}}

In this supplementary material, we present additional experimental results in \Cref{sec:sup-experiments}; additional ablation study in \Cref{sec:sup-ablation}; additional analyses in \Cref{sec:sup-analyses}.

\section{Additional Experiments}
\label{sec:sup-experiments}
Along with the evaluations on SR networks of scale $\times4$ discussed in the main manuscript, we also assess our framework on networks of scale $\times2$.
As shown in Table~\ref{tab:sup-scale2}, our framework outperforms existing SR quantization methods in terms of both PSNR and SSIM, demonstrating its effectiveness on scale $\times2$ SR networks.
Specifically, the PSNR gain on Set5 is 0.28 dB on EDSR, 0.74 dB on RDN, and 0.25 dB on SwinIR.
These results confirm that our framework is effective for both CNN- and Transformer-based SR networks of scale 2.
\begin{table}[ht]
\centering
\setlength{\tabcolsep}{1.2mm}
\caption{ \textbf{Quantitative comparisons on SR networks of scale $\times2$} 
}
\makebox[\linewidth]{\scriptsize
    \scalebox{1.25}{
    \begin{tabular}{l c cc cc cc cc}
        \toprule
        \multirow{2}{*}{Model} & \multirow{2}{*}{Bit} & 
        \multicolumn{2}{c}{Set5} & \multicolumn{2}{c}{Set14} & \multicolumn{2}{c}{B100} & \multicolumn{2}{c}{Urban100} \\
        \cmidrule(lr){3-4} \cmidrule(lr){5-6} \cmidrule(lr){7-8} \cmidrule(lr){9-10}& 
         & PSNR & SSIM & PSNR & SSIM & PSNR & SSIM & PSNR & SSIM \\
        \midrule
        EDSR   &32 & 37.93 & 0.960 & 33.46 & 0.916 & 32.10 & 0.899 & 31.71 & 0.925 \\
        \midrule
        EDSR-PAMS &2 &35.30&0.946 &31.63&0.899 &30.66&0.879 &28.11&0.875\\
        EDSR-DAQ  &2 &36.82&0.955 &32.50&0.908 &31.34&0.891 &29.85&0.905\\
        EDSR-DDTB &2 &37.25&0.958 &32.87&0.911 &31.67&0.893 &30.34&0.910\\
        EDSR-ODM (Ours)&2 &\textbf{37.53}& \textbf{0.958} & \textbf{33.06} & \textbf{0.913} & \textbf{31.81} & \textbf{0.895} & \textbf{30.81} & \textbf{0.915} \\
        \midrule
        RDN   &32 & 38.05 & 0.961 & 33.59 & 0.917 & 32.20 & 0.900 & 32.13 & 0.927 \\
        \midrule
        RDN-PAMS &2 &35.45&0.946 &31.67&0.899 &30.69&0.879 &28.14&0.874\\
        RDN-DAQ & 2 &37.23&0.957 &32.84&0.910 &31.66&0.893 &30.46&0.908\\
        RDN-DDTB &2 &36.76&0.955 &32.54&0.908 &31.44&0.890 &29.77&0.903\\
        RDN-ODM (Ours)&2 &\textbf{37.50}& \textbf{0.958} & \textbf{33.03} & \textbf{0.913} & \textbf{31.80} & \textbf{0.895} & \textbf{30.57} & \textbf{0.913} \\
        \midrule
        SwinIR  &32 & 38.14 & 0.961 & 33.86 & 0.921 & 32.31 & 0.901 & 32.76 & 0.934 \\
        \midrule
        SwinIR-PAMS &2 & 35.38 & 0.947 & 31.63 & 0.899 & 30.65 & 0.880 & 28.07 & 0.873\\
        SwinIR-DAQ  &2 & 34.98 & 0.943 & 31.38 & 0.896 & 30.47 & 0.876 & 27.83 & 0.869\\
        SwinIR-DDTB &2 & 37.17 & 0.957 & 32.78 & 0.911 & 31.42 & 0.888 & 30.24 & 0.908\\
        SwinIR-ODM (Ours)&2 &\textbf{37.42}& \textbf{0.958} & \textbf{33.03} & \textbf{0.913} & \textbf{31.79} & \textbf{0.895} & \textbf{30.76} & \textbf{0.914} \\
        \bottomrule
    \end{tabular}
}
}
\label{tab:sup-scale2}
\end{table}

Furthermore, we compare our method with existing methods by training each method for 300K iterations.
The results in \Cref{tab:sup-ep300} show that the gains achieved by our approach for 60K iterations reported in the main manuscript are maintained.
Our framework still achieves more than a 0.37 dB gain over other methods for Set5 when trained for extended iterations.
\begin{table}[ht]
\centering
\setlength{\tabcolsep}{1.2mm}
\caption{ \textbf{Quantitative comparisons of SR quantization methods with 300K iterations} 
}
\makebox[\linewidth]{\scriptsize
    \scalebox{1.25}{
    \begin{tabular}{l c cc cc cc cc}
        \toprule
        \multirow{2}{*}{Model} & \multirow{2}{*}{Bit} & 
        \multicolumn{2}{c}{Set5} & \multicolumn{2}{c}{Set14} & \multicolumn{2}{c}{B100} & \multicolumn{2}{c}{Urban100} \\
        \cmidrule(lr){3-4} \cmidrule(lr){5-6} \cmidrule(lr){7-8} \cmidrule(lr){9-10}& 
         & PSNR & SSIM & PSNR & SSIM & PSNR & SSIM & PSNR & SSIM \\
        \midrule
        EDSR   &32 & 32.10&0.894 &28.58&0.781 &27.56&0.736 &26.04&0.785\\
        \midrule
        EDSR-PAMS &2 &30.05&0.852 &27.17&0.744 &26.70&0.705 &24.09&0.707\\
        EDSR-DAQ &2 & 31.11&0.874 &27.98&0.765 &27.14&0.720 &24.96&0.745\\
        EDSR-DDTB &2 &31.19&0.878 &27.97&0.767 &27.14&0.723 &25.01&0.749\\
        EDSR-ODM (Ours)&2 &\textbf{31.56}& \textbf{0.884} & \textbf{28.15} & \textbf{0.771} & \textbf{27.30} & \textbf{0.728} & \textbf{25.24} & \textbf{0.758} \\
        \bottomrule
    \end{tabular}
}
}
\label{tab:sup-ep300}
\end{table}

Also, to ensure a fair comparison with DAQ~\cite{hong2022daq}, we follow their settings and apply our method to EDSR of 32 residual blocks with 256 channel dimensions.
As shown in \Cref{tab:sup-fulledsr}, our method outperforms DAQ even though DAQ employs a channel-wise quantization function, whereas our method utilizes a more efficient layer-wise function.
\begin{table}[ht]
\centering
\setlength{\tabcolsep}{1.2mm}
\caption{ \textbf{Quantitative comparisons on EDSR (full) of scale $\times4$} which consists of 32 residual blocks and 256 channel dimensions. For a fair comparison with DAQ, our model (ODM$^*$) is trained for 300K iterations.
}
\makebox[\linewidth]{\scriptsize
    \scalebox{1.25}{
    \begin{tabular}{l c cc cc cc cc}
        \toprule
        \multirow{2}{*}{Model} & \multirow{2}{*}{Bit} & 
        \multicolumn{2}{c}{Set5} & \multicolumn{2}{c}{Set14} & \multicolumn{2}{c}{B100} & \multicolumn{2}{c}{Urban100} \\
        \cmidrule(lr){3-4} \cmidrule(lr){5-6} \cmidrule(lr){7-8} \cmidrule(lr){9-10}& 
         & PSNR & SSIM & PSNR & SSIM & PSNR & SSIM & PSNR & SSIM \\
        \midrule
        EDSR-full & 32 & 32.46 & 0.897 & 28.80 & 0.788 & 27.72 & 0.742 & 26.64 & 0.803 \\
        \midrule
        EDSR-full-DAQ & 2 & 32.05 & 0.890 & 28.53 & 0.778 & 27.50 & 0.733 & 25.97 & 0.781 \\
        EDSR-full-ODM$^*$ (Ours)&2& \textbf{32.15} & \textbf{0.893} & \textbf{28.56} & \textbf{0.781} & \textbf{27.57} & \textbf{0.737} & \textbf{26.04} & \textbf{0.786}\\
        \bottomrule
    \end{tabular}
    }
}
\label{tab:sup-fulledsr}
\end{table}

Moreover, we compare our method with a fully-quantized SR network, EDSR-FQSR~\cite{Wang2021fully}, in which all layers and also the skip connections are quantized.
For a fair comparison, we also quantize all convolutional layers and the skip connections.
The results in \Cref{tab:sup-fully} show that our ODM outperforms FQSR, indicating that our approach is also effective when the network is fully quantized.
\begin{table}[!ht]
\centering
\caption{ \textbf{Quantitative comparisons on EDSR with fully quantized method.} 
S.C. refers to the bit-width of skip-connections.
}
\renewcommand{\arraystretch}{1.2}
\setlength{\tabcolsep}{0.8mm}
\makebox[\linewidth]{\scriptsize
    \scalebox{1.2}{
    \begin{tabular}{cl cc cc cc cc cc}
        \toprule
        \multirow{2}{*}{Scale}& \multirow{2}{*}{Model} & \multirow{2}{*}{Bit} & \multirow{2}{*}{S.C.} & 
        \multicolumn{2}{c}{Set5} & \multicolumn{2}{c}{Set14} & \multicolumn{2}{c}{B100} & \multicolumn{2}{c}{Urban100} \\
        \cmidrule(lr){5-6} \cmidrule(lr){7-8} \cmidrule(lr){9-10} \cmidrule(lr){11-12}& 
         & & & PSNR & SSIM & PSNR & SSIM & PSNR & SSIM & PSNR & SSIM \\
        \midrule
         & EDSR &32 & 32 & 32.10&0.894 &28.58&0.781 &27.56&0.736 &26.04&0.785\\
        \cmidrule(lr){2-12}
        $\times4$ & EDSR-FQSR & 4 & 8 & 30.93 & 0.870 & 27.82 & 0.761 & 27.07 & 0.715 & 24.93 & 0.744\\
        & EDSR-ODM (Ours) & 4 & 8 & \textbf{31.99} & \textbf{0.890} & \textbf{28.42} & \textbf{0.777} & \textbf{27.47} & \textbf{0.733} & \textbf{25.70} & \textbf{0.775}\\
        \midrule
        & EDSR &32 & 32 & 37.93 & 0.960 & 33.46 & 0.916 & 32.10 & 0.899 & 31.71 & 0.925\\
        \cmidrule(lr){2-12}
        $\times2$ & EDSR-FQSR & 4 & 8 & 37.04 & 0.951 & 32.84 & 0.908 & 31.67 & 0.889 & 30.65 & 0.911\\
        & EDSR-ODM (Ours) & 4 & 8 & \textbf{37.86} & \textbf{0.960} & \textbf{33.42} & \textbf{0.916} & \textbf{32.08} & \textbf{0.898} & \textbf{31.71} & \textbf{0.924}\\
        \bottomrule
    \end{tabular}
}}
\label{tab:sup-fully}
\end{table}

Additionally, along with SwinIR, as demonstrated in the main manuscript, we also apply our method to a more recent, large Transformer-based model, HAT~\cite{chen2023activating} ($\sim$10M parameters).
The results in Table~\ref{tab:sup-hat} indicate that our method can be effectively applied to Transformer models.
\begin{table}[ht]
\centering
\setlength{\tabcolsep}{1.2mm}
\caption{ \textbf{Quantitative comparisons on HAT of scale $\times{4}$} 
}
\makebox[\linewidth]{\scriptsize
    \scalebox{1.25}{
    \begin{tabular}{l c cc cc cc cc}
        \toprule
        \multirow{2}{*}{Model} & \multirow{2}{*}{Bit} & 
        \multicolumn{2}{c}{Set5} & \multicolumn{2}{c}{Set14} & \multicolumn{2}{c}{B100} & \multicolumn{2}{c}{Urban100} \\
        \cmidrule(lr){3-4} \cmidrule(lr){5-6} \cmidrule(lr){7-8} \cmidrule(lr){9-10}& 
         & PSNR & SSIM & PSNR & SSIM & PSNR & SSIM & PSNR & SSIM \\
        \midrule
        HAT   &32 & 32.92&0.905 &29.15 &0.796 &27.97&0.751 &27.87&0.835\\
        \midrule
        HAT-PAMS &2 &30.30&0.859 &27.35&0.750 &26.78&0.710 &24.23&0.713\\
        HAT-DAQ  &2 &30.21&0.856 &27.29&0.747 &26.74&0.708 &24.14&0.708\\
        HAT-DDTB &2 &31.23&0.878 &27.96&0.766 &27.16&0.724 &25.08&0.751\\
        HAT-ODM (Ours) &2 &\textbf{32.06}& \textbf{0.891} & \textbf{28.56} & \textbf{0.780} & \textbf{27.53} & \textbf{0.736} & \textbf{26.10} & \textbf{0.787} \\
        \bottomrule
    \end{tabular}
}
}
\label{tab:sup-hat}
\end{table}

\section{Additional Ablation Study}
\label{sec:sup-ablation}
In this section, we present an ablation study on the hyperparameters of our framework.
First, we conduct an ablation study on the percentile $j$, which is used to initialize quantization range clipping parameters.
The results in \Cref{tab:sup-ablation-p} show that when $j$=100, meaning the quantization range is not clipped and is determined by the maximum value, accuracy severely degrades.
This hints that clipping is important for performance and that using the max function does not serve as an effective initialization policy.

Also, we analyze the impact of the gradient balance terms $\lambda_R$ and $\lambda_M$, which balances the gradient of the reconstruction loss and the mismatch regularization loss, and initial learning rate $\beta^0$.
The results in \Cref{tab:sup-ablation-m,tab:sup-ablation-r} support our choice of $\lambda_R$=1, $\lambda_M$=1e-5, and $\beta^0$=1e-4.
\begin{table}[!ht]
\caption{
    \textbf{Ablation on hyperparameters} on EDSR $\times4$ (2-bit)
}
\centering
\setlength{\tabcolsep}{0.8mm}
    \newcommand{\h}{0.135\linewidth}
    \subfloat[Ablation on j\label{tab:sup-ablation-p}]{
        \resizebox*{!}{\h}{
            \begin{tabular}{lc ccc}
                \toprule
                $j$ & Set5 & Set14 & B100 & U100 \\
                \midrule
                100 & 29.63&26.86&26.53&23.82\\
                \textbf{99} & 31.50&28.14&27.27&25.17\\ 
                95 & 31.51&28.13&27.27&25.18\\ 
                \bottomrule
            \end{tabular}
        }
    }
    \subfloat[Ablation on $\lambda_M$ \label{tab:sup-ablation-r}]{
        \resizebox*{!}{\h}{
            \begin{tabular}{lc ccc}
                \toprule
                $\lambda_M$ & Set5 & Set14 & B100 & U100 \\
                \midrule
                2e-5 & 31.40 & 28.07 & 27.22 & 25.09\\
                \textbf{1e-5} & 31.50 & 28.14 & 27.27 & 25.17\\ 
                1e-6 & 31.48 & 28.15 & 27.28 & 25.18\\ 
                \bottomrule
            \end{tabular}
        }
    }
    \subfloat[Ablation on $\beta^0$ \label{tab:sup-ablation-m}]{
        \resizebox*{!}{\h}{
            \begin{tabular}{lc ccc}
                \toprule
                $\beta^0$ & Set5 & Set14 & B100 & U100 \\
                \midrule
                1e-3 & 30.82 & 27.72 & 27.01 & 24.63\\
                \textbf{1e-4} & 31.50 & 28.14 & 27.27 & 25.17\\ 
                1e-5 &  31.17 & 27.96 & 27.14 & 24.93\\ 
                \bottomrule
            \end{tabular}
        }
    }
\label{tab:sup-ablation}
\end{table}

Furthermore, we compare our weight clipping correction scheme with the commonly used quantization scheme for weights, LSQ~\cite{esser2019learned}.
The results in \Cref{tab:sup-ablation-lsq} validate the effectiveness of our clipping correction approach.
\begin{table}[!ht]
    \renewcommand{\arraystretch}{1.2}
    \setlength{\tabcolsep}{1.2mm}
    \centering
    \aboverulesep=0ex
    \belowrulesep=0ex
    \caption{\textbf{Comparison of WCC with LSQ} on EDSR $\times4$ (2-bit)
    }
    \resizebox{0.85\linewidth}{!}{
        \begin{tabular}{l| cc}
            \toprule
            Method & Set5 (PSNR/SSIM) & Urban100 (PSNR/SSIM) \\
            \midrule
            LSQ + Cooperative MR & 31.26 / 0.877 & 25.02 / 0.746\\
            WCC + Cooperative MR (Ours) & 31.50 / 0.882 & 25.17 / 0.755\\ 
            \bottomrule
        \end{tabular}
    }
    \label{tab:sup-ablation-lsq}
\end{table}

\section{Additional Analyses}
\label{sec:sup-analyses}
\subsection{Complexity Analysis}
\label{subsec:sup-complexity}
We provide additional complexity analysis on SwinIR in \Cref{tab:sup-complexity}.
The results show that our method achieves superior SR performance with minimal or no computational overhead in terms of model storage size and bitOPs.
BitOPs for SwinIR are calculated for processing a 64$\times$64 input patch.
\begin{table}[!ht]
    \centering
    \setlength{\tabcolsep}{1.4mm}
    \renewcommand{\arraystretch}{1.2}
    \aboverulesep=0ex
    \belowrulesep=0ex
    \caption{ 
        \textbf{Computational complexity comparison} on SwinIR of scale $\times4$
    }
    \resizebox{0.7\textwidth}{!}{
        \begin{tabular}{l|c|cc|cc}
            \toprule
            Model & Bit & Storage size & BitOPs & PSNR & SSIM \\
            \midrule
            SwinIR & 32 & 929.6K & 5.071T & 32.44 & 0.898 \\
            \midrule
            SwinIR-PAMS & 2 & 160.2K & 1.100T & 29.48 & 0.834\\
            SwinIR-DAQ  & 2 & 160.1K & 1.176T & 29.10 & 0.824 \\
            SwinIR-DDTB & 2 & 160.4K & 1.100T & 31.01 & 0.873 \\
            SwinIR-ODM (Ours) & 2 & \textbf{160.4K} & \textbf{1.100T} & \textbf{31.44} & \textbf{0.880} \\
            \bottomrule
        \end{tabular}
    }
\label{tab:sup-complexity}    
\end{table}

\subsection{Distribution mismatch}
For the generalizability of the observed distribution mismatch problem of the main manuscript, we analyze the variance of features in SR networks across a set of images.
As reported in \Cref{tab:sup-analyses-mismatch}, the classification network (ResNet20) exhibits much less image-wise and channel-wise variance compared to SR networks (EDSR, RDN).
This suggests that the distribution mismatch problem is particularly severe in SR networks. 
\begin{table}[!ht]
    \setlength{\tabcolsep}{1.4mm}
    \caption{
    {\textbf{Average variance in feature.}} The metrics are measured on DIV2K validation set for SR networks and ImageNet validation set for the classification network.} 
    \centering
    \resizebox{0.85\linewidth}{!}{
            \begin{tabular}{llc ccc}
                \toprule
                Task & Model & Image-wise Variance & Channel-wise Variance \\
                \midrule
                Image super-resolution & EDSR ($\times4$) & 15.08 & 40.29\\
                Image super-resolution & RDN ($\times4$) & 6.40 & 58.14 \\
                Image classification & ResNet-20 & 0.04 & 0.09\\
                \bottomrule
            \end{tabular}
    }
    \label{tab:sup-analyses-mismatch}
\end{table}

Moreover, we analyze the feature mismatch after quantization-aware training (QAT) using different methods.
To track feature similarity, we compute the variance between feature distribution statistics (mean and standard deviation) within the benchmark dataset.
As shown in \Cref{tab:sup-analyses-mismatch-reduction}, the feature mismatch is effectively reduced with our framework.
\begin{table}[!ht]
    \renewcommand{\arraystretch}{1.2}
    \setlength{\tabcolsep}{1.4mm}
    \centering
    \aboverulesep=0ex
    \belowrulesep=0ex
    \caption{
        {\textbf{Feature mismatch after quantization-aware training.}} The metrics are measured on DIV2K validation set for SR networks and ImageNet validation set for the classification network.
    } 
    \resizebox{0.65\textwidth}{!}{
    \begin{tabular}{l|ccc}
        \toprule
        Similarity metric & Var[mean] & Var[std] & Avg. Mismatch\\
        \midrule
        Before QAT & 0.28 & 3.60 & 2.41e+02 \\
        \midrule
        After QAT w/ \textbf{ODM} & \textbf{0.25} & \textbf{1.02} & \textbf{1.37e+02}\\
        After QAT w/ PAMS & 0.29 & 2.34 & 4.04e+02 \\
        After QAT w/ DDTB & 2.54 & 1.67 & 5.46e+05\\
        After QAT w/ DAQ & 2.14 & 6.59  & 6.10e+06\\
        \bottomrule
    \end{tabular}
    }
    \label{tab:sup-analyses-mismatch-reduction}
\end{table}

We provide additional analysis supporting the choice of distance from the quantization grid as a measure of distribution mismatch in \cref{eq:mismatch} of the main manuscript.
We note that QAT is a process searching for a quantization grid that best fits the discrepant input distributions. 
If the average distance of each feature from the quantization grid is small, it implies that most features are aligned with the grid, indicating a low distribution mismatch.
According to \Cref{tab:sup-analyses-mismatch-reduction}, it is verified that using distance as the mismatch measure reduces the variance in feature distribution statistics across test images.

\subsection{Cooperative Regularization}
\label{subsec:sup-cooperative}
In the main manuscript, we emphasized the importance of \textit{cooperatively} using mismatch regularization and reconstruction losses.
For the cooperative update, we weigh the gradient of mismatch regularization using the cosine similarity with the gradient of the reconstruction loss.
The weighing term is formulated as $0.5 \cdot (cos(\vv_a, \vv_b)+1)$.
Here, we present results using the weighing term as $cos(\vv_a, \vv_b)$ following Du~\etal~\cite{du2018adapting} and that of $u(cos(\vv_a, \vv_b))$ where $u(\cdot)$ is the unit step function.
The results in \Cref{tab:sup-cooperative} show that all these functions that alleviate the conflict between the two losses achieve high reconstruction accuracy, indicating that the general cooperative property is the key to performance gain.
\begin{table}[!ht]
    \renewcommand{\arraystretch}{1.2}
    \setlength{\tabcolsep}{1.2mm}
    \centering
    \aboverulesep=0ex
    \belowrulesep=0ex
    \caption{\textbf{Different formulation for cooperative behaviour} on EDSR $\times4$ (2-bit)
    }
    \resizebox{0.85\linewidth}{!}{
        \begin{tabular}{l| cc}
            \toprule
            Formulation & Set5 (PSNR/SSIM) & Urban100 (PSNR/SSIM) \\
            \midrule
            $cos(\vv_a, \vv_b)$~\cite{du2018adapting} & 31.46 / 0.881 & 25.14 / 0.755 \\
            $u(cos(\vv_a, \vv_b))$ & 31.49 / 0.883 & 25.15 / 0.756\\
            $0.5 \cdot (cos(\vv_a, \vv_b)+1)$ (Ours) & 31.50 / 0.882 & 25.17 / 0.755\\ 
            \bottomrule
        \end{tabular}
    }
    \label{tab:sup-cooperative}
\end{table}

\subsection{More Visualizations}
\label{subsec:sup-visualizations}
For better comprehension, we provide additional results of the effect of our loss after training in \Cref{fig:sup-dist}.
Along with the results in \Cref{fig:exp-dist} of the main manuscript, these results show that our loss term updates the activation distributions to a further quantization-friendly state while mostly preserving the high-density values of the original distribution.
\newcommand{\sidecaption}[2][\empty]
{\bgroup
  \sbox0{#2}
  \rotatebox[origin=Bl]{90}{\parbox{\ht0}{\subcaption[position=above]{#1}}}%
  \usebox0
\egroup}
\begin{figure}[!ht]
\centering
    \setlength{\tabcolsep}{0.05cm}
    \newcommand{\w}{0.22\linewidth}
    \begin{tabular}{cccc}
        \centering
        \sidecaption[Layer-6]{\includegraphics[width=\w]{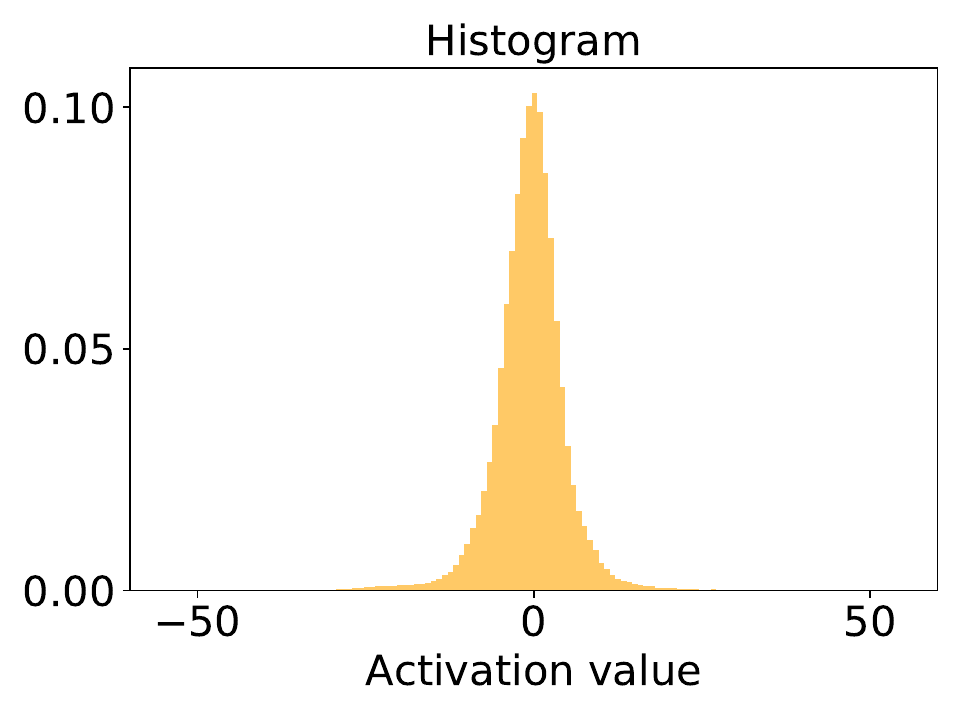}} &
        \includegraphics[width=\w]{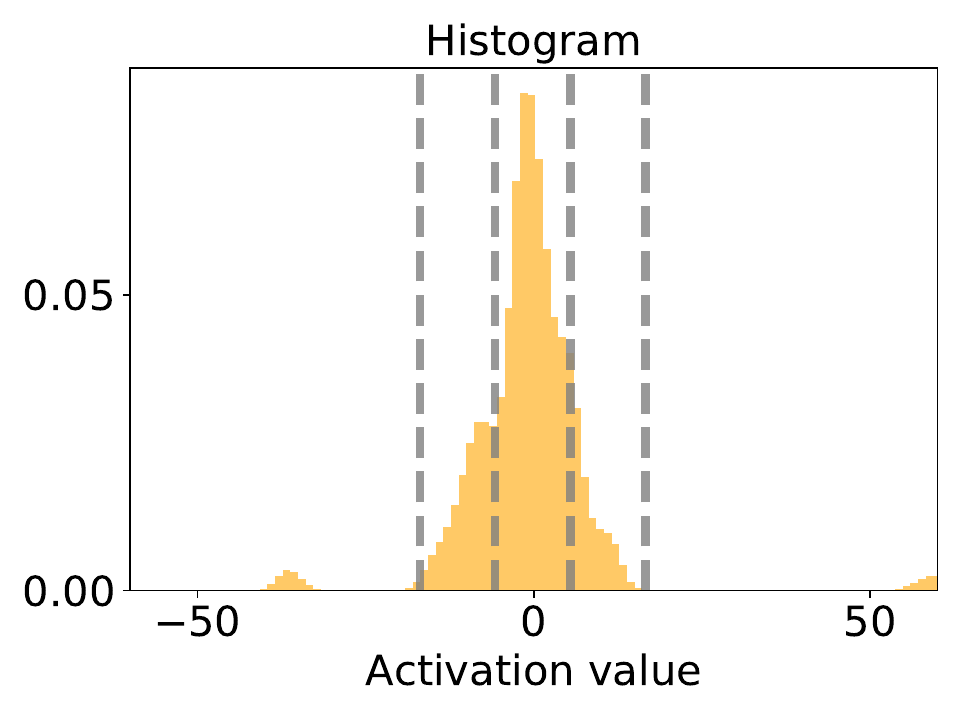} &
        \includegraphics[width=\w]{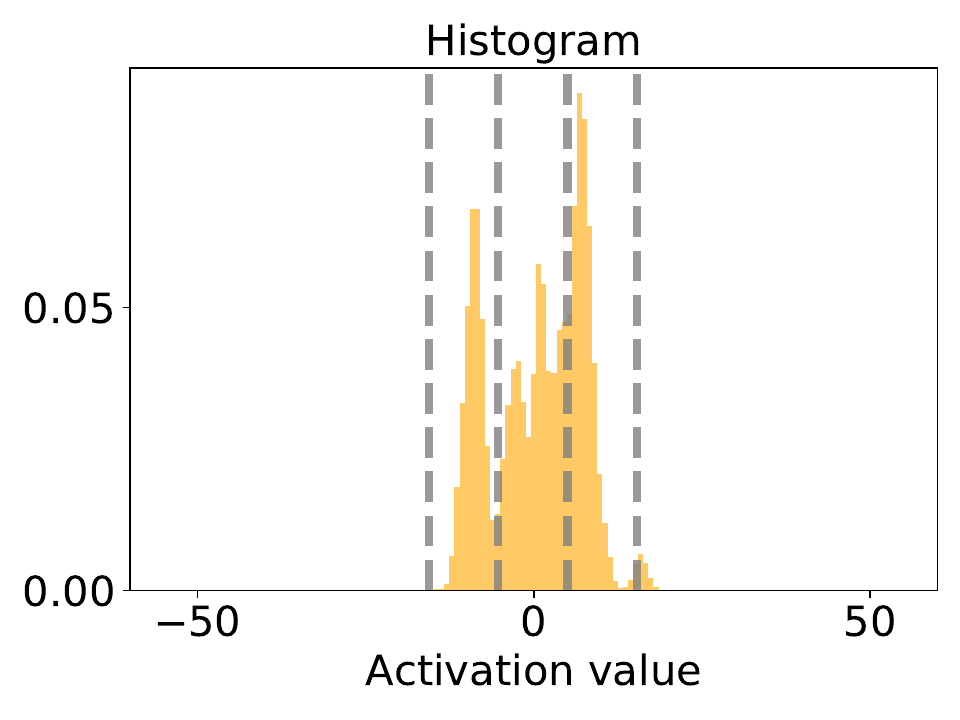} &
        \includegraphics[width=\w]{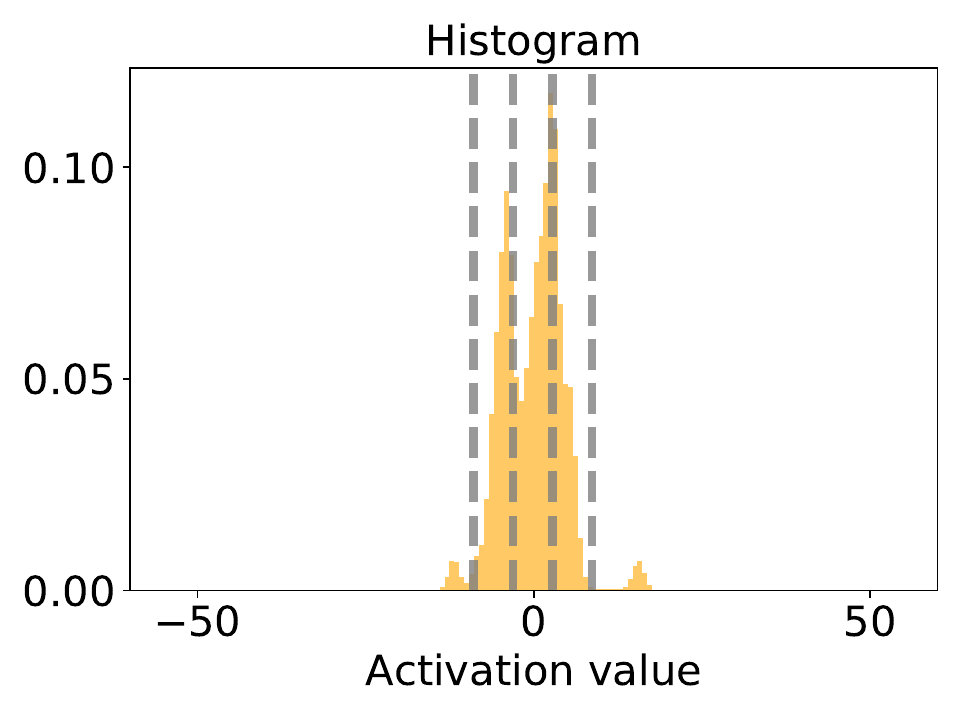} \\
        \sidecaption[Layer-12]{\includegraphics[width=\w]{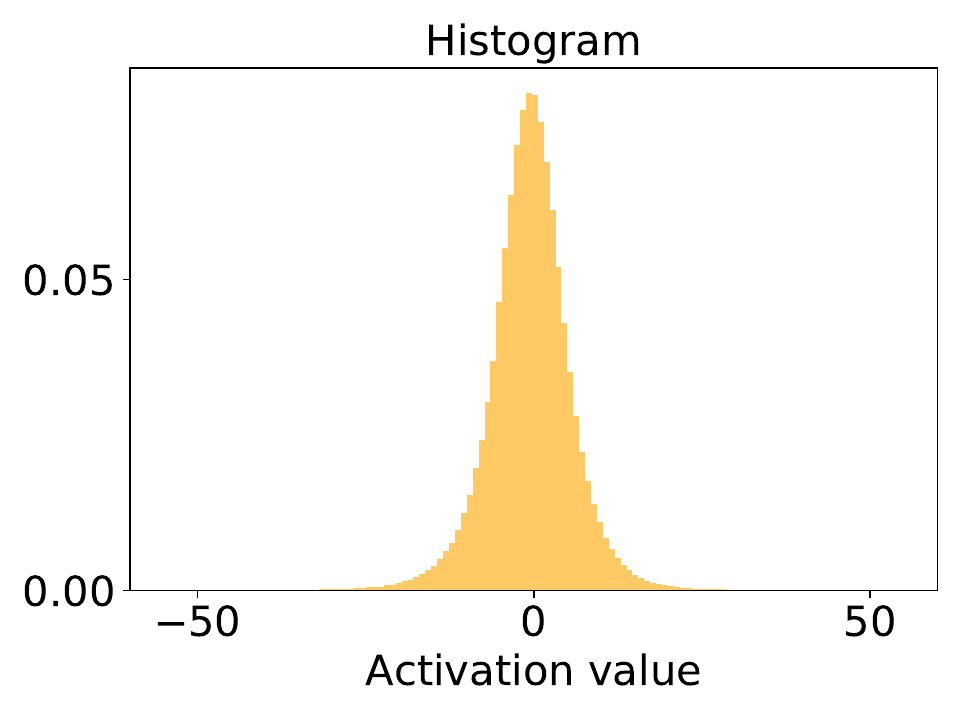}} &
        \includegraphics[width=\w]{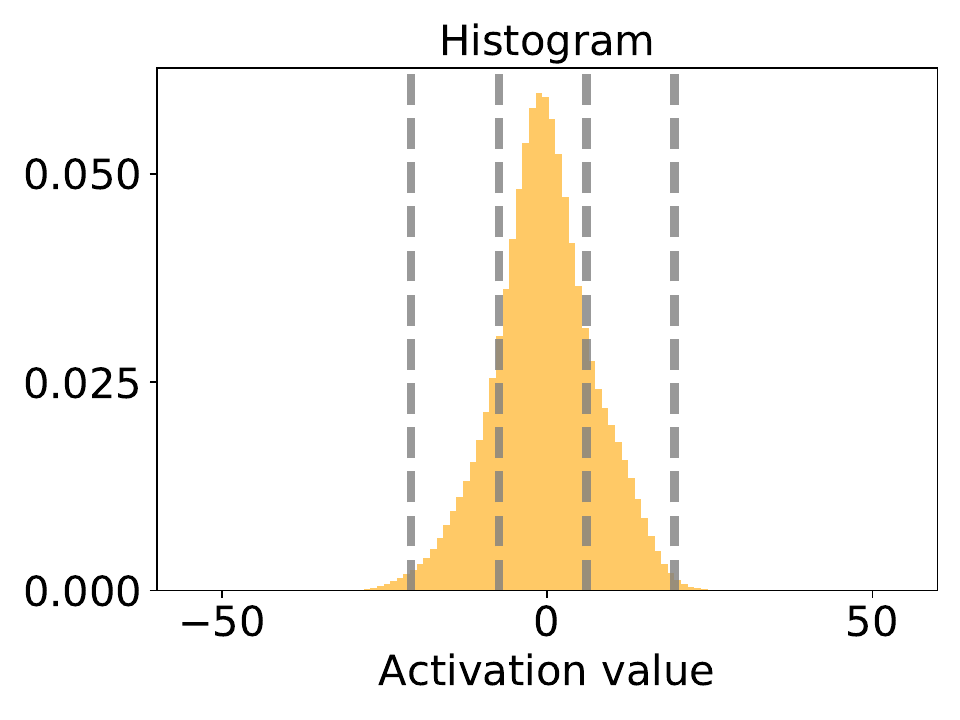} &
        \includegraphics[width=\w]{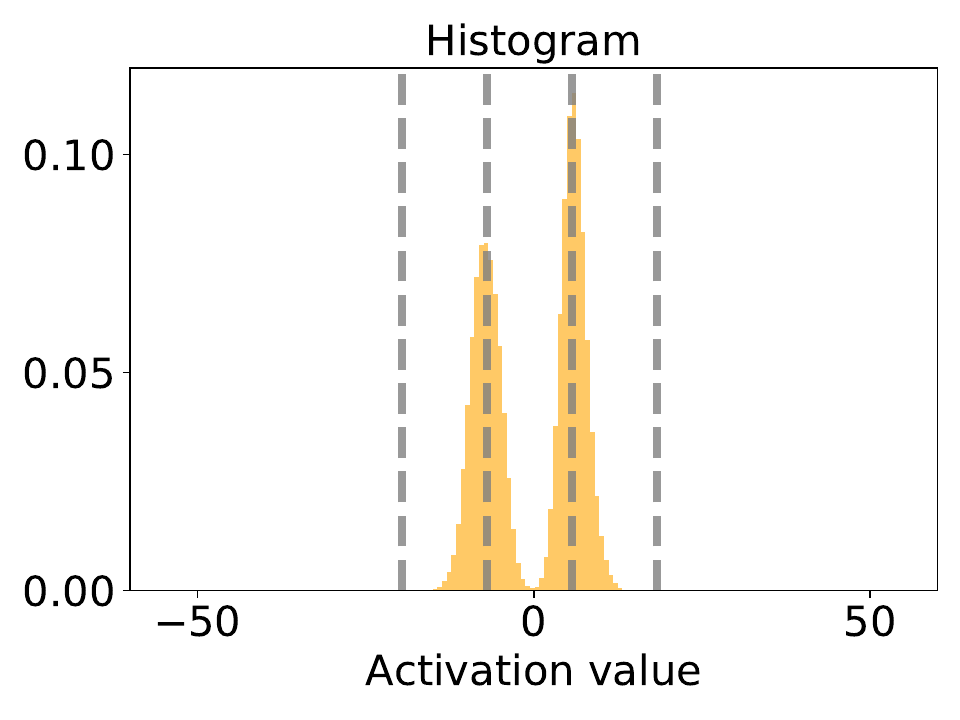} &
        \includegraphics[width=\w]{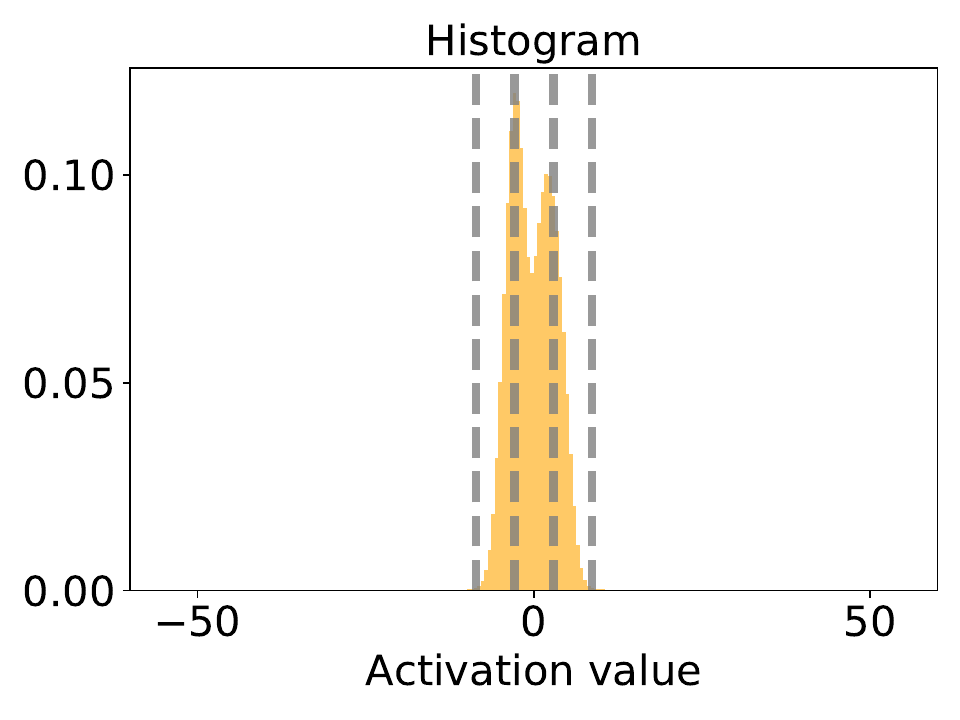} \\
        \sidecaption[Layer-16]{\includegraphics[width=\w]{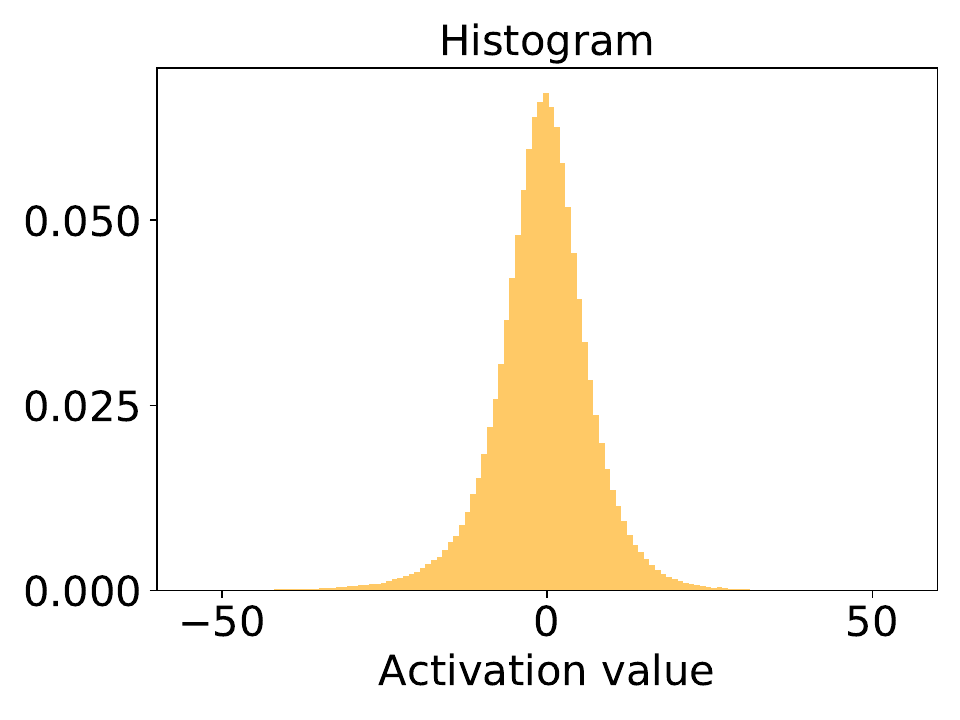}} &
        \includegraphics[width=\w]{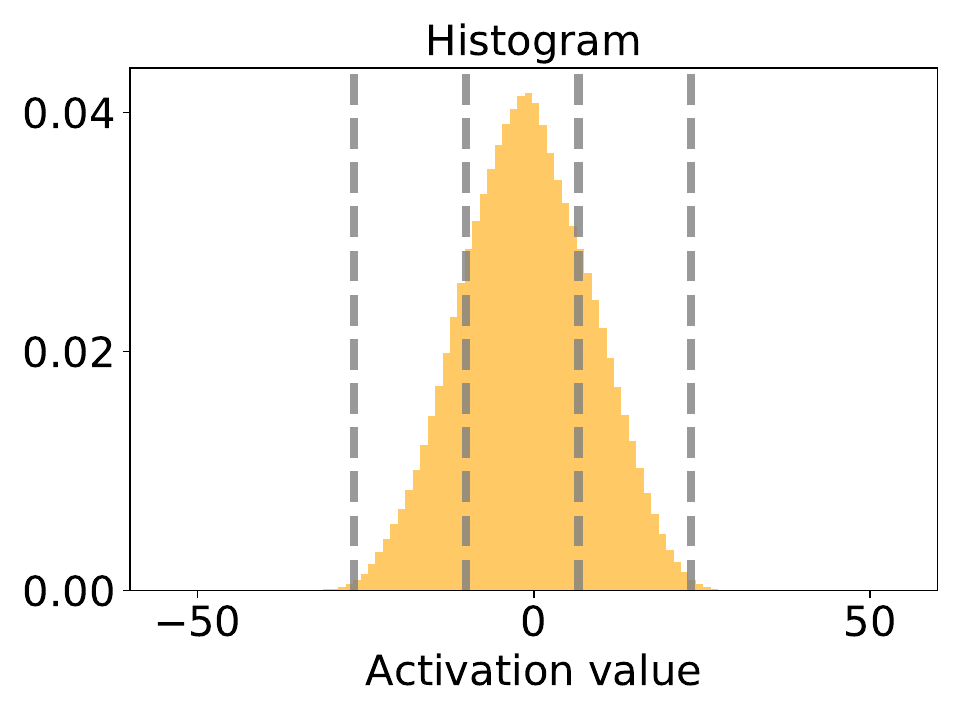} &
        \includegraphics[width=\w]{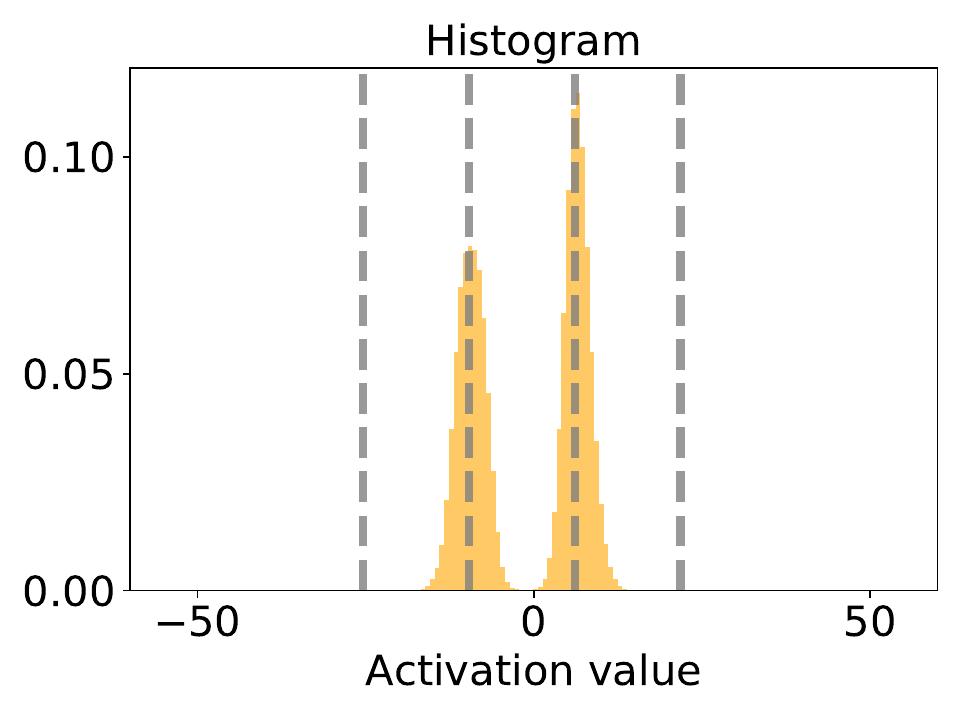} &
        \includegraphics[width=\w]{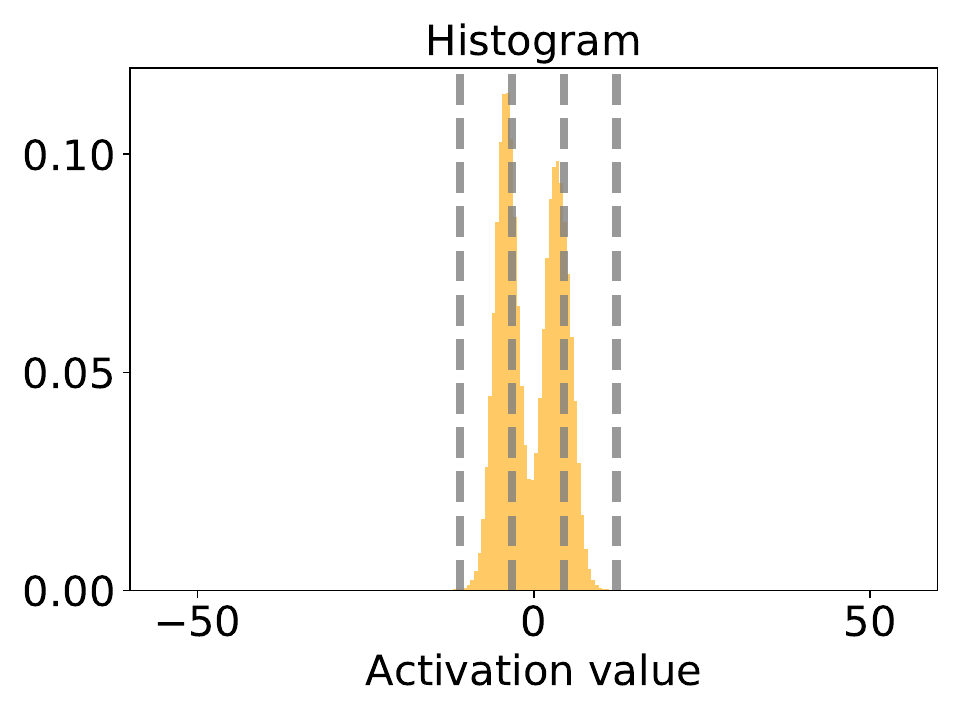} \\
        \sidecaption[Layer-20]{\includegraphics[width=\w]{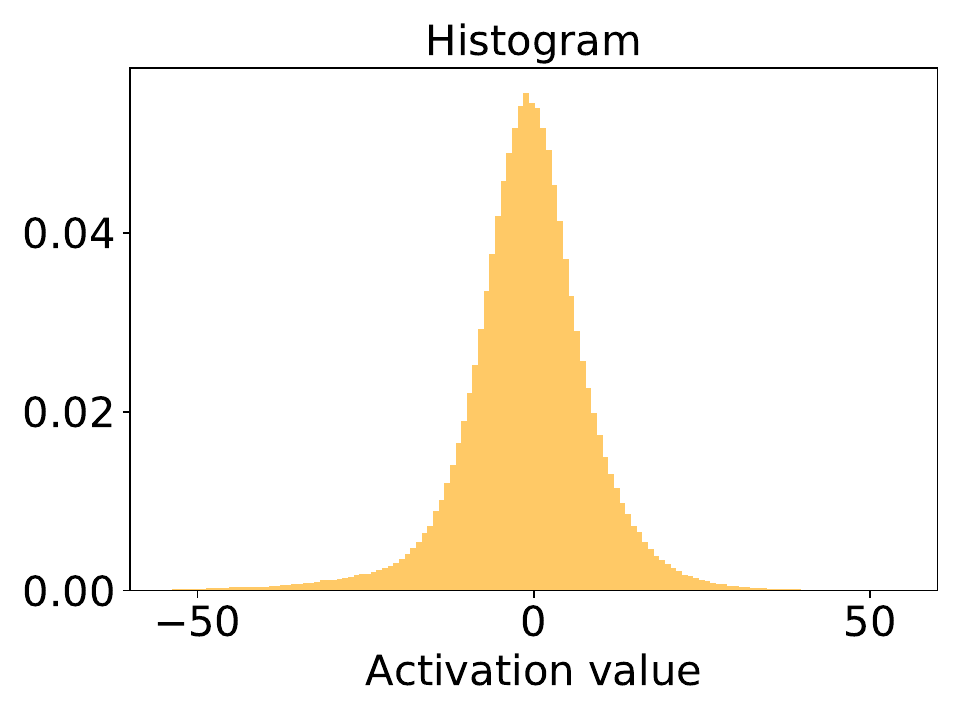}} &
        \includegraphics[width=\w]{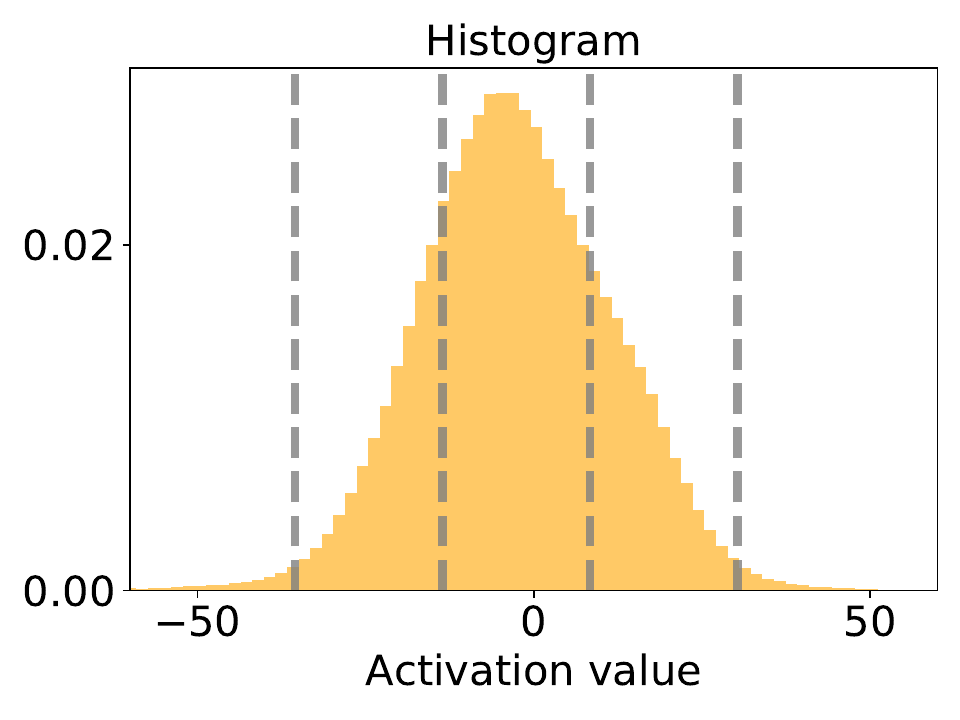} &
        \includegraphics[width=\w]{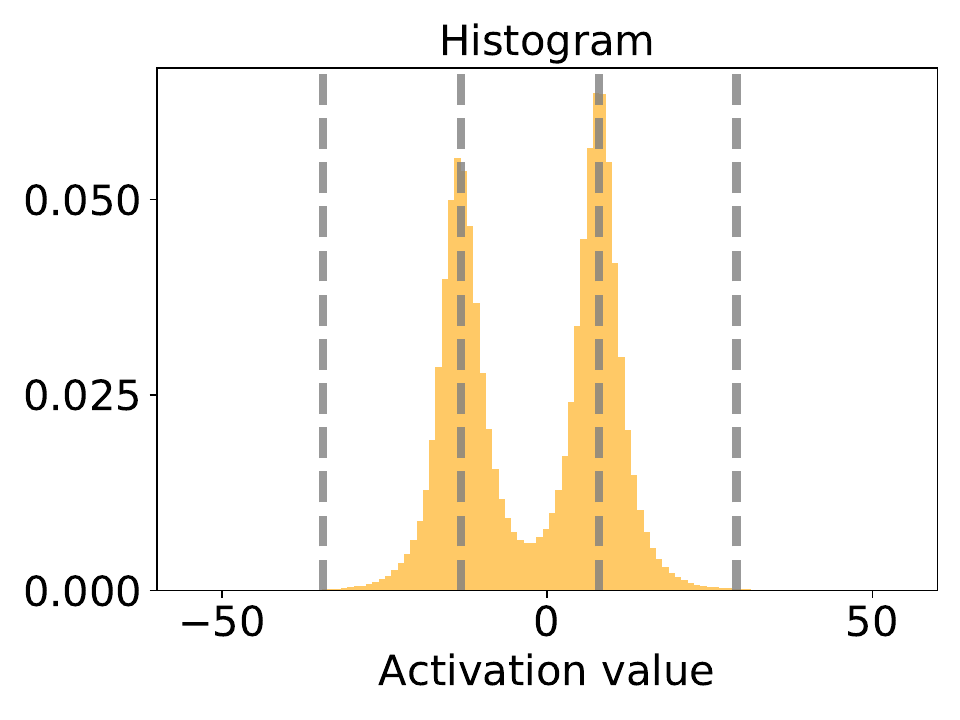} &
        \includegraphics[width=\w]{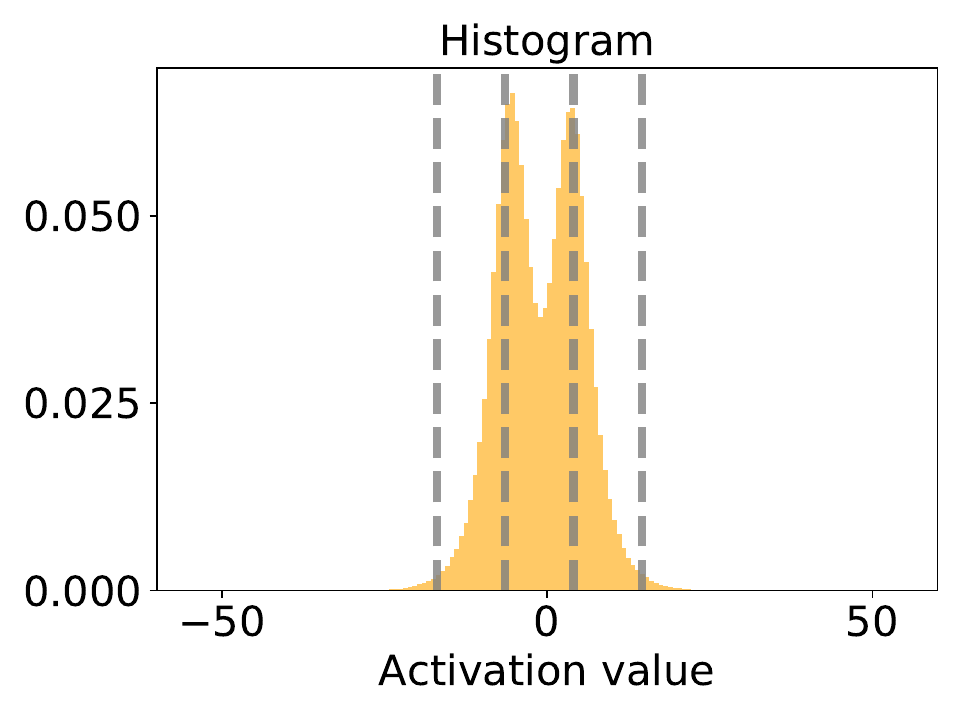} \\
        \scriptsize{32-bit network} & \scriptsize{$\mathcal{L}_{R}$} & \scriptsize{$\mathcal{L}_{R}+\mathcal{L}_{M}$} & \scriptsize{Cooperative MR} \\
    \end{tabular}
    \caption{
    \textbf{Distribution of activations after training and before quantization.}
    Activations of the different convolution layers in 2-bit EDSR-ODM are visualized.
    }
    \label{fig:sup-dist}
\end{figure}

\begin{figure}[!ht]
    \centering
    \begin{subfigure}{0.32\textwidth}
        \centering
        \setlength{\tabcolsep}{1.2mm}
        \includegraphics[width=0.95\textwidth]{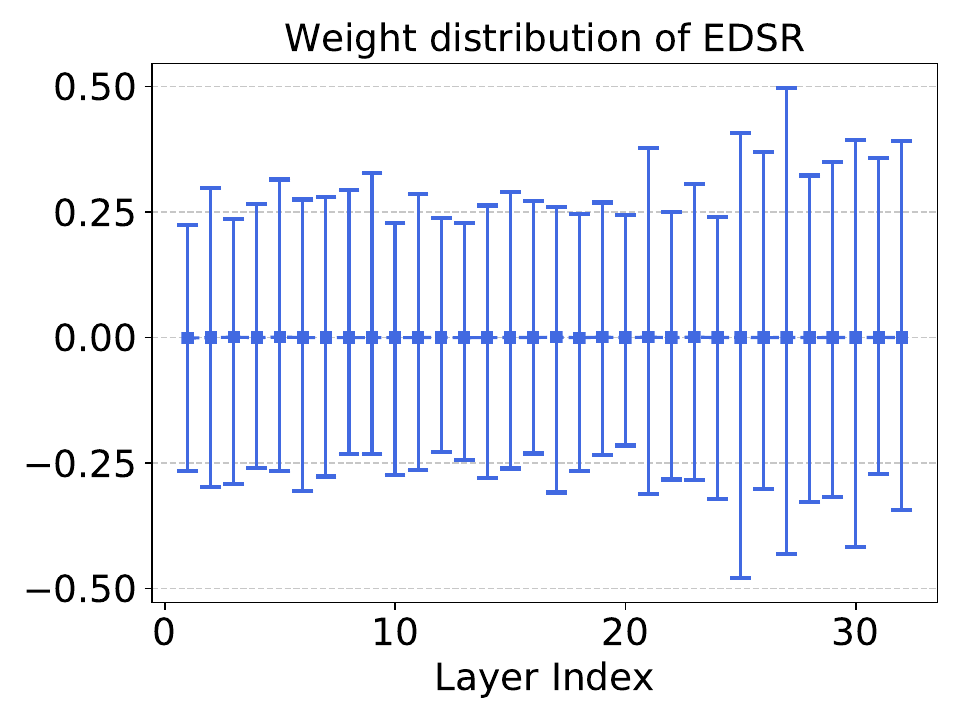}
        \caption{EDSR\label{fig:sup-weight-a}}
    \end{subfigure}
    \begin{subfigure}{0.32\textwidth}
        \centering
        \setlength{\tabcolsep}{1.2mm}
        \includegraphics[width=0.95\textwidth]{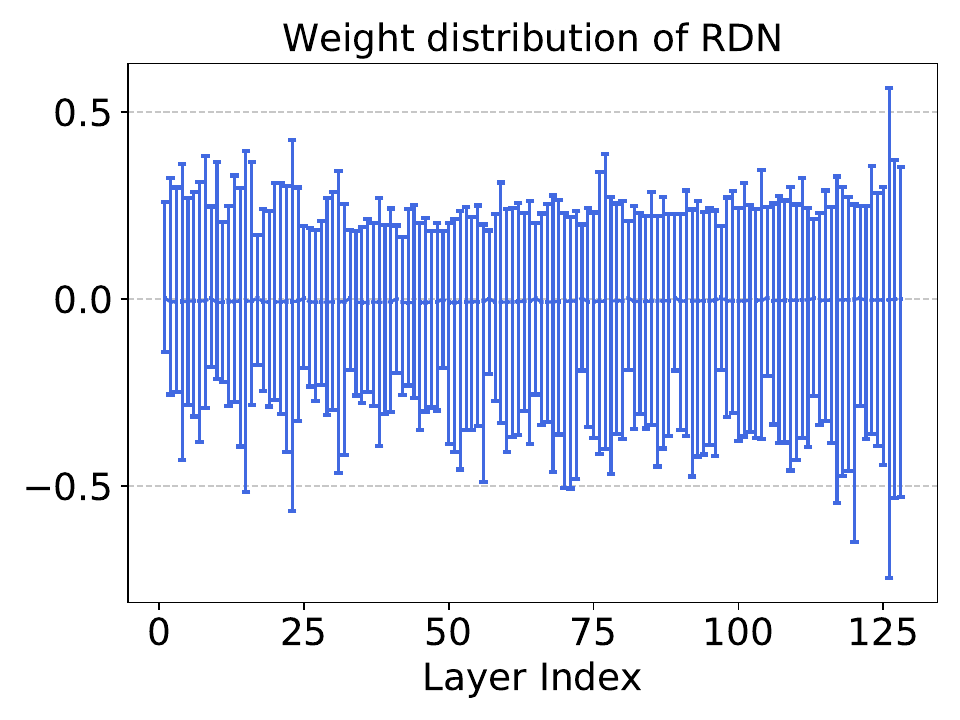}
        \caption{RDN\label{fig:sup-weight-b}}
    \end{subfigure} 
    \begin{subfigure}{0.32\textwidth}
        \centering
        \setlength{\tabcolsep}{1.2mm}
        \includegraphics[width=0.95\textwidth]{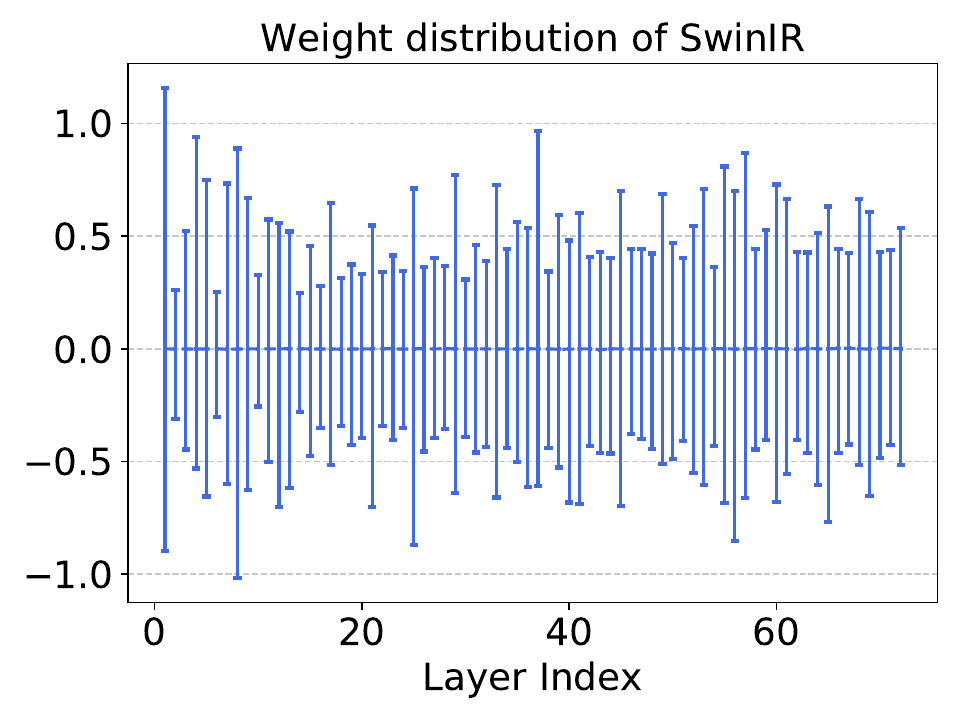}
        \caption{SwinIR\label{fig:sup-weight-c}}
    \end{subfigure} 
    \caption{ 
        \textbf{Layer-wise different weight distributions of SR networks.} 
        The minimum, maximum, and mean of each convolution/linear layer weight are plotted.
    }
    \label{fig:sup-weight}
\end{figure}

Moreover, we provide visualizations of layer-wise mismatch in weights for different SR networks in \Cref{fig:sup-weight}.
According to the visual results, the weight distributions of different layers have a similar mean (\ie, near 0), but exhibit varying minimum and maximum values.
This motivates us to use a layer-wise different policy for determining the weight quantization range.

\subsection{Training Time}
\label{subsec:sup-traintime}

Although our framework primarily aims to achieve an accurate quantized SR network in which the inference cost is reduced via quantization, we also provide comparisons on the training time.
According to \Cref{tab:sup-traintime}, our training scheme requires a shorter training time than DDTB and DAQ.
Although our training incurs slightly more time overhead compared to PAMS, the gains in test accuracy compensate for this additional training cost.
\begin{table}[!ht]
    \centering
    \setlength{\tabcolsep}{1.2mm}
    \caption{
    \textbf{Training time} of QAT methods on SR networks. The training time is measured by running the experiment on a single RTX 2080Ti GPU.
    } 
    \resizebox{0.8\linewidth}{!}{
        \begin{tabular}{l cccc}
            \toprule
            Method & EDSR-PAMS & EDSR-DAQ & EDSR-DDTB & EDSR-ODM (Ours) \\
            \midrule
            Time (hours) & 1.5 & 3.7 & 2.5 & 2.4 \\ 
            \bottomrule
        \end{tabular}
    }
    \label{tab:sup-traintime}
\end{table}

\subsection*{License of the Used Assets}
\begin{itemize}
    \item[$\bullet$] DIV2K~\cite{agustsson2017ntire}  dataset is publicly available for academic research purposes.
    \item[$\bullet$] Set5~\cite{bevilacqua2012low}, Set14~\cite{ledig2017photo}, BSD100~\cite{martin2001database}, Urban100~\cite{huang2015single} datasets are made available at \href{https://github.com/jbhuang0604/SelfExSR}{https://github.com/jbhuang0604/SelfExSR}.
\end{itemize}

\end{document}